\documentclass[a4paper,fleqn]{cas-dc}
\usepackage{tikz}
\usetikzlibrary{mindmap,trees,shapes.geometric}
\usepackage{longtable}
\usepackage{hyperref}
\usepackage{orcidlink}
\usetikzlibrary{positioning, arrows.meta, shapes, backgrounds}
\usepackage[x11names]{xcolor}
\usepackage[colorinlistoftodos,prependcaption,textsize=small]{todonotes}
\usepackage[most]{tcolorbox}
\usepackage{tabularx}
\usepackage{booktabs}
\usepackage{threeparttable}

\definecolor{MyBlue}{RGB}{26,13,171}

\usepackage[authoryear]{natbib}

\def\tsc#1{\csdef{#1}{\textsc{\lowercase{#1}}\xspace}}
\tsc{}
\tsc{}
\begin{document}
\let\WriteBookmarks\relax
\def\floatpagepagefraction{1}
\def\textpagefraction{.001}

\shorttitle{}    

\shortauthors{}  

\title [mode = title]{Agentic AI with Orchestrator-Agent Trust: A Modular Visual Classification Framework with Trust-Aware Orchestration and RAG-Based Reasoning}


\tnotetext[2]{The publication of the article in OA mode was financially supported by HEAL-Link. This work was additionally supported in part by the National Science Foundation (NSF) and the United States Department of Agriculture (USDA), National Institute of Food and Agriculture (NIFA), through the “Artificial Intelligence (AI) Institute for Agriculture” program under Award Numbers AWD003473 and AWD004595, and USDA-NIFA Accession Number 1029004 for the project titled “Robotic Blossom Thinning with Soft Manipulators.” Additional support was provided through USDA-NIFA Grant Number 2024-67022-41788, Accession Number 1031712, under the project “Expanding UCF AI Research To Novel Agricultural Engineering Applications (PARTNER).} 


%[<options>]

\author[1]{Konstantinos I. Roumeliotis}[orcid=0000-0002-8098-1616]
\cormark[1]
\ead{k.roumeliotis@uop.gr}
\affiliation[1]{organization={University of the Peloponnese, Department of Informatics and Telecommunications},city={Tripoli},postcode={22131},country={Greece}}

\author[2]{Ranjan Sapkota}[orcid=0000-0002-5417-6744]
\cormark[1]
\ead{rs2672@cornell.edu}
\affiliation[2]{organization={Cornell University, Department of Biological and Environmental Engineering},city={Ithaca},postcode={14850},state={NY},country={USA}}

\author[2]{Manoj Karkee}[orcid=0000-0001-5337-4848]
\ead{mk2684@cornell.edu}
\cormark[1]

\author[1]{Nikolaos D. Tselikas}[orcid=0000-0001-5799-3558]
\ead{ntsel@uop.gr}
\cormark[1]

\cortext[1]{Corresponding authors}



\begin{abstract}
Modern Artificial Intelligence (AI) increasingly relies on multi-agent architectures that blend visual and language understanding. Yet, a pressing challenge remains: How can we trust these agents especially in zero-shot settings with no fine-tuning? We introduce a novel modular Agentic AI visual classification framework that integrates generalist multimodal agents with a non-visual reasoning orchestrator and a Retrieval-Augmented Generation (RAG)  module. Applied to apple leaf disease diagnosis, we benchmark three configurations: (I) zero-shot with confidence-based orchestration, (II) fine-tuned agents with improved performance, and (III) trust-calibrated orchestration enhanced by CLIP-based image retrieval and re-evaluation loops. Using confidence calibration metrics (ECE, OCR, CCC), the orchestrator modulates trust across agents. Our results demonstrate a 77.94\% accuracy improvement in the zero-shot setting using trust-aware orchestration and RAG, achieving 85.63\% overall. GPT-4o showed better calibration, while Qwen-2.5-VL displayed overconfidence. Furthermore, image-RAG grounded predictions with visually similar cases, enabling correction of agent overconfidence via iterative re-evaluation. The proposed system separates perception (vision agents) from meta-reasoning (orchestrator), enabling scalable and interpretable multi-agent AI. This blueprint illustrates how Agentic AI can deliver trustworthy, modular, and transparent reasoning, and is extensible to diagnostics, biology, and other trust-critical domains. In doing so, we highlight Agentic AI not just as an architecture but as a paradigm for building reliable multi-agent intelligence. All models, prompts, results, and system components including the complete software source code are openly released to support reproducibility, transparency, and community benchmarking at our \href{https://github.com/Applied-AI-Research-Lab/Orchestrator-Agent-Trust}{Github} page.
\end{abstract}


% \begin{highlights}
% \item 
% \item 
% \item 
% \end{highlights}

%\nocite{*}

% Keywords

\maketitle
\scriptsize
% \tableofcontents
\normalsize
% Main text
\section{Introduction}
\label{sec:Introduction}
The integration of vision and language models into autonomous decision-making systems has redefined the boundaries of artificial intelligence (AI), especially in fields that demand both perceptual accuracy and interpretability \cite{afroogh2024trust, ilievski2025human}. Multimodal large language models (LMMs), capable of reasoning over both visual and textual data, are increasingly employed in diverse domains, ranging from autonomous robotics and medical imaging to scientific diagnostics and agricultural monitoring \cite{bradshaw2025large, yang2024limits, mon2025embodied}. However, as these models are deployed in high-stakes environments, a fundamental challenge persists: can these systems be trusted to make reliable, transparent, and justifiable decisions particularly in zero-shot or open-world scenarios where no prior task-specific fine-tuning is possible?

Recent developments in agentic AI systems where autonomous agents collaborate, reason, and interact with their environment have highlighted the importance of meta-reasoning and modularity in AI architectures \cite{ale2025enhancing, buehler2025preflexor, sapkota2025ai}. Rather than relying on monolithic end-to-end networks, agentic AI systems distribute cognitive tasks across specialized AI agents \cite{savaglio2020agent}. In vision-language applications, this paradigm enables distinct agents to independently process visual inputs, generate explanations, and assess confidence, while an external orchestrator or supervisor performs higher-order reasoning to synthesize their outputs \cite{jeyakumar2024advancing, zhai2024fine, lin2025showui}. This architectural shift not only mirrors aspects of human collaborative problem-solving but also introduces an additional layer of oversight, which is critical for ensuring accountability and trust \cite{qiao2025oversight}.

Despite the promise of this approach, trust calibration within agentic AI remains an underexplored area. Conventional ensemble systems often assume that agents' self-reported confidence scores are reliable proxies for correctness \cite{ma2024you, warmsley2025self, wang2025impact, huang2024llm}, an assumption that breaks down in zero-shot generalist models such as GPT-4o or Qwen-2.5-VL. These models may exhibit systematic overconfidence or fail to discriminate between subtle categories in domain-specific tasks \cite{wen2024mitigating}. Consequently, there is a pressing need to assess and correct misalignments between an agent’s confidence and its actual performance. This problem is particularly acute in scientific and diagnostic domains, where misclassification may lead to erroneous conclusions or interventions.

In visual classification tasks such as plant disease detection, trustworthiness is not merely a matter of predictive accuracy but of interpretability and justification \cite{nigar2024improving, ding2024next, qadri2024advances}. For instance, two models may arrive at the same prediction but differ significantly in the rationale behind their decision \cite{ali2023explainable, bajorath2025scientific}. A reliable AI system should not only be accurate but also capable of articulating why a decision was made and when to defer to alternative evidence \cite{messeri2024artificial}.  RAG  approaches where models consult external databases or prior examples to refine their outputs offer a promising pathway to ground predictions in visual context, enhancing both trust and transparency \cite{ke2025retrieval, dong2025understand, tozuka2025application}.

Deploying generalist VLMs in high-stakes, domain-specific scenarios presents critical challenges in calibration, interpretability, and decision trustworthiness. In agricultural automation and digital diagnostics, for instance, misclassifications can lead to economic losses and delayed interventions. Traditional deep learning pipelines offer high accuracy but lack modular reasoning and trust introspection.

\begin{figure*}[t]
\centering
\begin{tcolorbox}[
    colback=blue!5!white,
    colframe=blue!95!black,
    title=Three-Stage Trust-Aware Agentic AI Framework,
    fonttitle=\bfseries,
    boxrule=0.8pt,
    arc=3pt,
    left=2pt,
    right=2pt,
    top=4pt,
    bottom=4pt,
    width=\textwidth % span both columns
]

\centering
\resizebox{\textwidth}{!}{% auto-scale to full page width
\begin{tikzpicture}[node distance=1.8cm and 2.2cm,
    every node/.style={font=\scriptsize\sffamily}]

    % Main experiment boxes
    \node (exp1) [draw, rounded corners, fill=red!15,
        minimum width=2.6cm, minimum height=1.2cm, align=center] 
        {\textbf{Experiment I}\\Zero-Shot Agents\\\textit{+ Confidence-Aware}\\\textit{Orchestration}};
        
    \node (exp2) [right=of exp1, draw, rounded corners, fill=orange!15,
        minimum width=2.6cm, minimum height=1.2cm, align=center] 
        {\textbf{Experiment II}\\Fine-Tuned Agents\\\textit{+ Confidence-Aware}\\\textit{Orchestration}};
        
    \node (exp3) [right=of exp2, draw, rounded corners, fill=green!15,
        minimum width=3.0cm, minimum height=1.5cm, align=center] 
        {\textbf{Experiment III}\\Zero-Shot Agents\\\textit{+ Trust-Aware Orchestration}\\\textit{+ RAG + Re-Evaluation Loop}};
    
    % Arrows between stages
    \draw[->, thick] (exp1) -- (exp2)
        node[midway, above, yshift=2pt] {\tiny \textbf{Agent Fine-Tuning}};
    \draw[->, thick] (exp2) -- (exp3)
        node[midway, above, yshift=2pt] {\tiny \textbf{Add Trust Metrics + RAG}};

    % Bottom annotations
    \node[below=1.2cm of exp1] {\tiny \textbf{Generalist Multimodal Models}};
    \node[below=1.2cm of exp2] {\tiny \textbf{Domain-Specialized via Supervised Fine-Tuning}};
    \node[below=1.2cm of exp3] {\tiny \textbf{Trust-Calibrated + Evidence-Grounded Pipeline}};

    % Model names below each box
    \node[below=0.3cm of exp1] {\tiny GPT-4o, Qwen-2.5-VL (Zero-Shot)};
    \node[below=0.3cm of exp2] {\tiny GPT-4o, Qwen-2.5-VL (Fine-Tuned)};
    \node[below=0.3cm of exp3] {\tiny GPT-4o, Qwen-2.5-VL + CLIP (RAG)};

\end{tikzpicture}
}% end resizebox
\end{tcolorbox}
\caption{Three-stage trust-aware agentic AI framework across experiments.}
\label{fig:three-stage-trust-aware-agentic-ai}
\end{figure*}
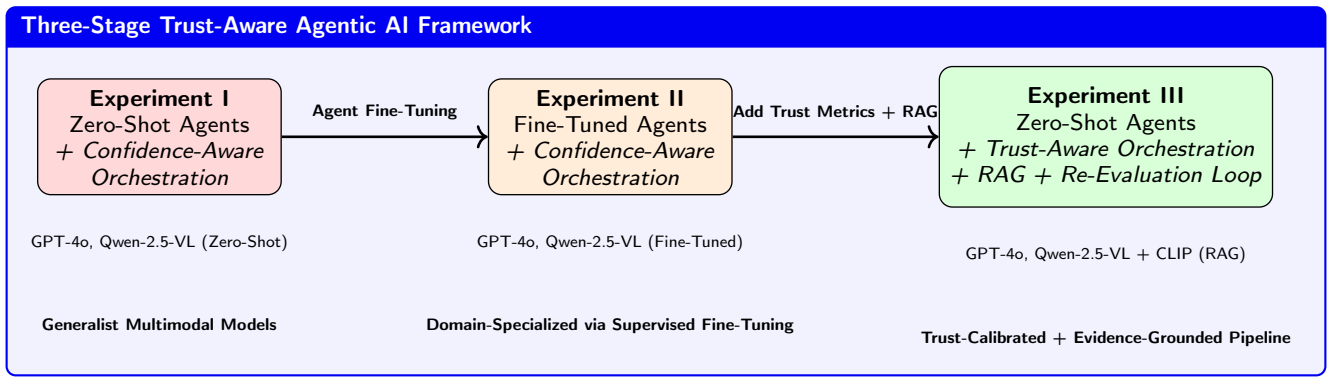

To address these limitations, we propose a novel Agentic AI framework for visual classification that integrates trust-aware orchestration with retrieval-augmented reasoning in a modular, interpretable, and zero-shot-capable architecture. Our system coordinates multiple multimodal agents GPT-4o and Qwen-2.5-VL with a non-visual orchestrator that synthesizes final predictions based on reported confidence, natural language justifications, and internal trust scores. When discrepancies or low-confidence predictions arise, the orchestrator triggers a re-evaluation loop using  RAG  supported by CLIP-based image retrieval, allowing the system to reflect and refine its decision using similar visual precedents. We introduce a three-stage pipeline (see Figure~\ref{fig:three-stage-trust-aware-agentic-ai}), incrementally augmenting the system’s capabilities.

The overall flow diagram and key contributions of this study are illustrated in Figure~\ref{fig:intro}, highlighting how our trust-aware Agentic AI framework enables scalable integration of vision-language agents, dynamic trust calibration, and retrieval-augmented reasoning. Each experimental stage builds on the previous to increase reasoning fidelity and trust interpretability:

\begin{figure*}[!htbp]
\centering
\includegraphics[width=\textwidth]{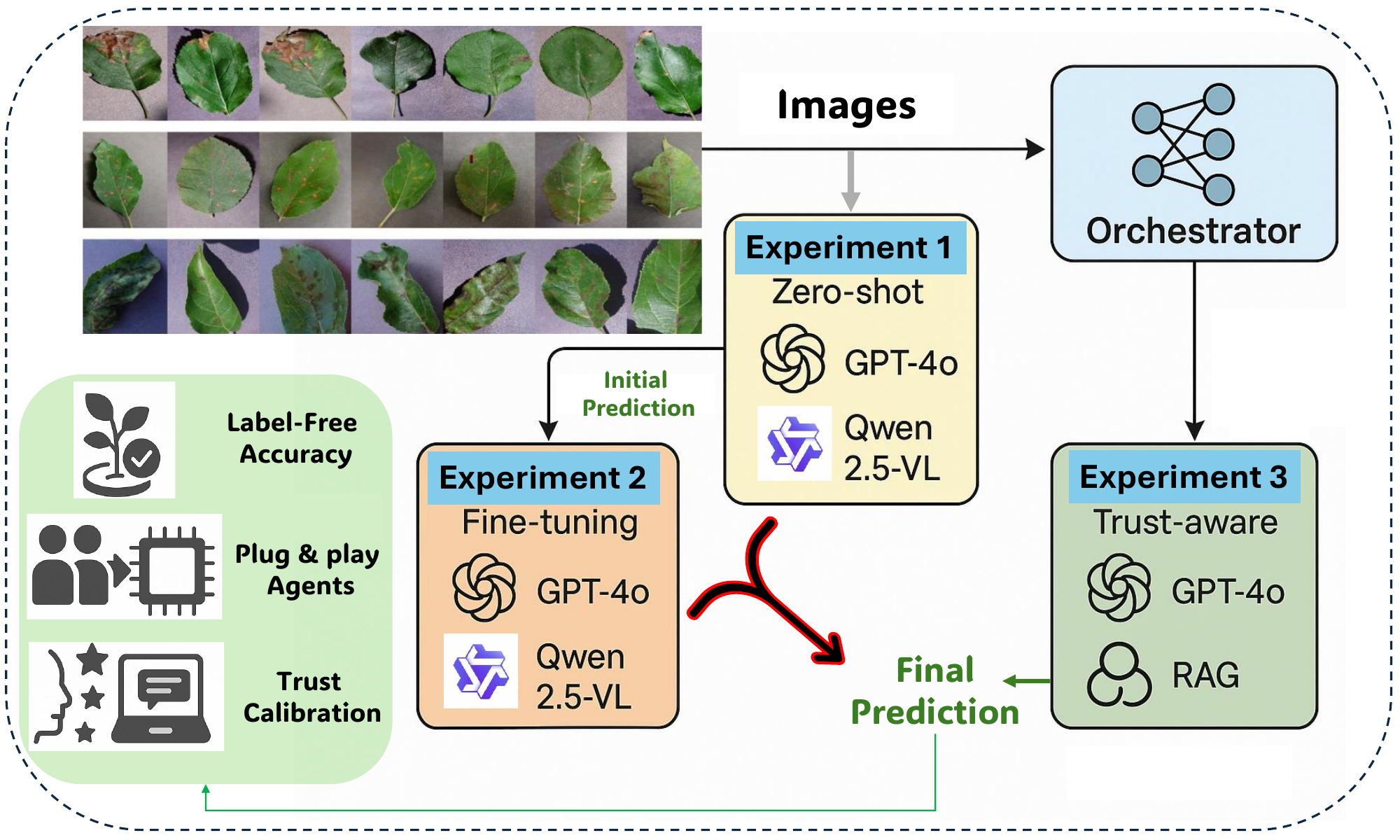}
\caption{Overview of our trust-aware Agentic AI framework for visual classification, illustrating modular agents, orchestration stages, trust calibration, and retrieval-augmented re-evaluation for accurate, interpretable decisions. \textbf{This workflow is designed for plant leaf disease classification but is generalizable to any RGB image classification task involving multimodal agents and trust-aware decision pipelines.}}
\label{fig:intro}
\end{figure*}

\begin{itemize}
    \item \textbf{Experiment I:} AI Agents operate in zero-shot mode; the orchestrator selects final predictions using reported confidence scores.
    \item \textbf{Experiment II:} AI Agents are fine-tuned on a curated apple leaf disease dataset using hyperparameter optimization (Bayesian search) to evaluate the effects of supervised domain adaptation.
    \item \textbf{Experiment III:} Image-RAG, trust-aware orchestration, and agent re-evaluation are integrated into a unified Agentic AI system. A trust evaluation layer quantifies agent reliability using metrics such as Expected Calibration Error (ECE), Overconfidence Ratio (OCR), and Consistency Gap (CG), enabling the orchestrator to make informed arbitration decisions. The same general-purpose multimodal agents introduced in Experiment I independently classify images with natural language rationales and confidence scores; when trust metrics signal unreliability, the orchestrator initiates a re-evaluation loop that supplies agents with prior context and Image-RAG retrieval results. Agents then either revise or reaffirm their predictions, and the orchestrator makes a final decision based on these updated responses.
\end{itemize}

We validate this framework on a biologically grounded task: the classification of apple leaf diseases, where fine-grained distinctions (e.g., between rust and scab) and explanation interpretability are critical. Our results show a relative improvement of 77.94\% in classification accuracy from 48.13\% in the baseline to 85.63\% in the trust-aware configuration even without additional fine-tuning in Experiment III. This demonstrates the power of structured trust arbitration and retrieval-enhanced reasoning for improving performance and interpretability in real-world, open-domain applications.

In summary, our contributions are fourfold: (1) A modular agentic AI system that decouples perception, reasoning, and retrieval; (2) a novel trust-aware orchestration strategy grounded in multi-dimensional calibration; (3) a CLIP-RAG-based re-evaluation loop for uncertainty mitigation; and (4) a comprehensive empirical validation across three reasoning regimes with reproducible and scalable design.

\section{Methodology}
\label{Methods}
This study proposes and evaluates a modular Agentic AI architecture for visual classification, combining two vision-language agents (GPT-4o and Qwen-2.5-VL) with a non-visual orchestrator (o3-mini-2025-01-31). Experiments were conducted using a curated dataset of 800 RGB images labeled into four apple disease categories, split 64\% for training, 16\% for validation, and 20\% for testing.

The system comprises three experimental setups.

\subsection{Experiment I}
\label{Experiment_I}
In the first experimental configuration, we evaluate the performance of general-purpose multimodal LLMs in a zero-shot setting that is, without any task-specific fine-tuning.

Each agent receives a prompt containing a single input image (from the test set) and is tasked with predicting the most appropriate plant disease class. In addition to the predicted class label, agents are required to return (i) a confidence score in the range [0.0–1.0] and (ii) a natural language explanation justifying their decision (Appendix, Fig.~\ref{fig:appendix_agent_prompt}). These three outputs classification, confidence, and rationale are compiled and forwarded to a non-visual orchestrator model (o3-mini-2025-01-31 \cite{OpenAIInc2025b}), which serves as a comparative reasoner. The orchestrator does not process images directly; instead, it evaluates the agents' predictions and justifications in light of their associated confidence scores and produces a final classification decision through structured, confidence-aware reasoning (Appendix, Fig.~\ref{fig:appendix_orchestrator_prompt}).

\subsection{Experiment II}
\label{Experiment_II}
In Experiment II, both agents were fine-tuned using supervised learning techniques to improve classification performance. For GPT-4o, fine-tuning employed a hyperparameter configuration informed by prior ResNet-50 optimization studies. Qwen-2.5-VL underwent over 50 hyperparameter tuning trials, beginning with heuristic parameter estimates and refined through performance-based search strategies.
\begin{enumerate}
\item \textbf{GPT-4o Fine-Tuning:} In this phase, we explored two fine-tuning strategies: one using default hyperparameters provided by the OpenAI platform, and one informed by prior hyperparameter optimization conducted on a ResNet-50 model using Bayesian optimization.
\begin{enumerate}
    \item ResNet-50–Informed Hyperparameter Transfer. To mitigate the computational cost of performing hyperparameter optimization directly on GPT-4o, we hypothesized that high-performing hyperparameters derived from ResNet-50 tuning on the same dataset could be effectively transferred to GPT-4o. Specifically, Bayesian optimization with the Tree-structured Parzen Estimator (TPE) algorithm was used to explore the ResNet-50 hyperparameter space across 30 trials. TPE iteratively models the objective function \( f(x) \), evaluates the expected improvement (EI) of candidate configurations, and selects promising trials using a likelihood ratio.
    
    Optimization was implemented using the Optuna library on an NVIDIA A100-SXM4-40GB GPU via Colab Enterprise. Early stopping and pruning were used to improve computational efficiency. The best configuration identified 10 training epochs and a batch size of 16 was subsequently applied to GPT-4o fine-tuning through the OpenAI fine-tuning interface. Training and validation sets were formatted into \texttt{jsonl} files with prompt–completion pairs before submission.
    \item Default Hyperparameter Configuration. In parallel, we conducted fine-tuning using the default hyperparameters recommended by the OpenAI platform: 3 epochs and a batch size of 1. Identical training and validation files were used to ensure a fair comparison.
\end{enumerate}
Fine-tuning GPT-4o using the ResNet-50–informed hyperparameters required approximately 1,778 seconds (\(\sim\)29.6 minutes) and incurred a cost of USD~47.53. In comparison, fine-tuning with the default hyperparameters (3 epochs, batch size of 1) required 1,652 seconds (\(\sim\)27.5 minutes) at a reduced cost of USD~13.09. Although direct hyperparameter optimization on GPT-4o could potentially yield higher-performing configurations, the computational and financial cost of conducting such a process over multiple trials renders it impractical under current constraints.

A detailed comparison of runtime, cost, and validation loss for both configurations is provided in the Appendix, Table~\ref{tab:app_exp2_finetune}.

Based on these results, the ResNet-50–informed configuration yielded a substantially lower validation loss (0.0088) and was therefore selected as the preferred GPT-4o variant for integration into the Agentic AI system implemented in this study.
\item \textbf{Qwen-2.5-VL-7B Fine-Tuning:} To adapt the Qwen-2.5 Vision-Language 7B (VL-7B) model for our task, we fine-tuned the \texttt{unsloth/Qwen2.5-VL-7B-Instruct-bnb-4bit} model \cite{Wang2024}, a 4-bit quantized variant of the original pretrained Qwen2.5 model. This quantized version enables efficient loading and inference with reduced memory requirements, making it suitable for commodity GPU hardware. We employed Low-Rank Adaptation (LoRA) in 16-bit precision to inject trainable adapters into the model, thereby allowing effective fine-tuning without full dequantization of the backbone weights.

The LoRA adapters were integrated into all major architectural components of the model, including the vision encoder layers, the language modeling transformer layers, multi-head attention modules, and the feedforward multilayer perceptrons (MLPs). Each adapter was configured with a rank \( r = 16 \) and scaling factor \( \alpha = 16 \), balancing parameter efficiency with expressive capacity. We used the Parameter-Efficient Fine-Tuning (PEFT) framework to update only the LoRA-injected weights, keeping the underlying 4-bit quantized parameters frozen throughout training.

To identify the optimal hyperparameter configuration, we employed Bayesian optimization using the Optuna framework, leveraging the TPE as the sampler. The search space included learning rate, per-device batch size, gradient accumulation steps, warmup ratio, weight decay, and number of training epochs. Optimization targeted the minimization of validation loss. Trial results were persistently stored in both \texttt{jsonl} and SQLite formats, allowing for checkpointing and resumability in the event of interruption.

Poor-performing trials were automatically pruned using Optuna's \texttt{MedianPruner}, which compares intermediate results to the median of previous completed trials and terminates underperformers early. Early stopping was also employed during training to prevent overfitting and reduce unnecessary computational expenditure. All training was conducted using the \texttt{SFTTrainer} class from the Hugging Face \texttt{trl} library, which supports supervised fine-tuning with periodic evaluation at the end of each epoch. Throughout training, only the LoRA adapters were updated, while the quantized backbone remained frozen to preserve efficiency.

Recognizing the potential inefficiency of starting hyperparameter optimization from purely random initial conditions, particularly in high-dimensional or sensitive parameter spaces, we manually enqueued a strong initial configuration to guide the search. This warm-start configuration used a learning rate of \(2 \times 10^{-4}\), a per-device batch size of 2, 8 gradient accumulation steps, a warmup ratio of 0.05, a weight decay of 0.01, and 10 training epochs. This guided initialization enabled the optimizer to begin exploring in a region of the parameter space known to yield promising results.

A total of 50 hyperparameter optimization trials were performed. Among these, 20 trials completed successfully, while 30 were pruned based on intermediate performance. No trials failed due to runtime errors. The best-performing trial achieved a validation loss of \(1.0 \times 10^{-5}\), using a learning rate of approximately \(1.094 \times 10^{-5}\), a batch size of 4, 4 gradient accumulation steps, a warmup ratio of 0.0997, a weight decay of 0.00127, and 15 training epochs (Appendix, Table~\ref{tab:top_5_hyperparameter_configurations}).
\end{enumerate}

During inference, both the fine-tuned GPT-4o and Qwen agents received a single input image and generated three outputs: (i) a predicted disease category, (ii) a natural language explanation justifying the classification, and (iii) a normalized confidence score in the range \([0, 1]\). As in the previous configuration (Experiment I), a non-visual orchestrator model (\texttt{o3-mini-2025-01-31}) served as the final decision-maker. This orchestrator performed comparative reasoning by evaluating structured prompts that included each agent’s predictions, explanations, and confidence scores. Based on this synthesis, it produced a consolidated classification accompanied by a rationale for its selection.

\subsection{Experiment III}
\label{Experiment_III}
In the final experimental configuration, we integrate in our Agentic AI system CLIP-based image retrieval (Image-RAG), trust-aware orchestration, and agent re-evaluation mechanisms. The orchestrator incorporates similarity-based retrieval results and calibration metrics to determine whether to trust agent outputs or trigger re-evaluation loops. The system dynamically guides agents to revise their predictions when confidence or justification is misaligned with expected trust signals (Fig.~\ref{fig:appendix_exp3_rag}).

\textbf{Trust Evaluation Layer.}
The Trust Evaluation Layer conducts offline trust profiling of the AI agents by quantifying their calibration, discriminative reliability, and consistency over a labeled dataset. Specifically, both agents were evaluated in a zero-shot inference setting on the training image set originally employed in Experiment \hyperref[Experiment_II]{II} for fine-tuning, but here presented without ground-truth labels during evaluation. For each image, we recorded the predicted class, associated confidence score, and explanatory output provided by the agent. Using this comprehensive log of model behavior, we computed a suite of quantitative trustworthiness metrics to assess each agent’s predictive confidence alignment and overall reliability under zero-shot conditions.

\begin{itemize}
    \item \textbf{Expected Calibration Error (ECE):} In this study, ECE quantifies the discrepancy between a model's predicted confidence and its actual accuracy across prediction bins, serving as a key metric for assessing trustworthiness in zero-shot and fine-tuned agentic AI systems. A low ECE value indicates that the model's self-reported confidence aligns closely with its empirical correctness, enabling more reliable orchestration decisions based on agent trustworthiness. By incorporating ECE into the orchestrator's arbitration logic, we enhance the system's ability to identify overconfident failures and trigger re-evaluation loops when agent confidence is not well-calibrated.
    \[
    \text{ECE} = \sum_{m=1}^{M} \frac{|B_m|}{n} \left| \text{acc}(B_m) - \text{conf}(B_m) \right|,
    \]
    Where $B_m$ is the $m$-th confidence bin, $n$ the number of samples, and $\text{acc}$ and $\text{conf}$ represent per-bin accuracy and confidence.
    
    \item \textbf{Overconfidence Ratio (OCR):} The OCR measures the proportion of incorrect predictions made with high confidence (e.g., confidence > 0.9), capturing the extent to which models exhibit unjustified certainty. A high OCR value signals a critical misalignment between model confidence and actual correctness, often leading to erroneous decisions in confidence-based orchestration pipelines. In our trust-aware framework, OCR serves as a penalizing factor in agent arbitration, allowing the orchestrator to detect and downweight predictions from agents that frequently make confidently wrong decisions.
    
    \[
    \text{OCR} = \frac{|\{i : \hat{y}_i \neq y_i \land c_i > 0.9\}|}{|\{i : c_i > 0.9\}|},
    \]
    Measuring how often models are confidently wrong.
    \item \textbf{Consistency Gap (CG):} The CG quantifies the divergence in a model’s predictions when exposed to semantically equivalent but syntactically varied prompts, serving as a proxy for reasoning stability under linguistic perturbations. A large CG indicates that the AI agent's output is highly sensitive to prompt phrasing, undermining reliability and reproducibility in decision-making. Within our framework, CG is used as a trust signal to evaluate the robustness of agent responses, enabling the orchestrator to identify models prone to inconsistency and initiate re-evaluation when necessary.
    \[
    \text{CG} = \frac{1}{n} \sum_{i=1}^{n} \mathbb{I}[P_i^{(1)} \neq P_i^{(2)}],
    \]
    Assessing variation in outputs across prompt formulations.
\end{itemize}

This quantitative profiling enabled agent-level trust estimation: a method for assessing an agent’s general reliability independently of individual inputs. These trust profiles were then used by the orchestrator to modulate the influence of each agent in decision fusion granting higher weight to agents with stronger calibration and lower overconfidence.

\textbf{Image RAG.}
To complement the Agentic AI classification framework, we incorporated a multimodal retrieval-augmented generation (Image-RAG) system designed for plant disease detection via semantic similarity and weighted voting. This framework utilizes a pre-trained vision-language model (CLIP, ViT-B/32 variant) to embed input images into a high-dimensional feature space that supports interpretable and context-rich decision-making.

Given an input image $\mathbf{I} \in \mathbb{R}^{H \times W \times 3}$, the CLIP vision encoder transforms it into a 512-dimensional embedding $\mathbf{e} \in \mathbb{R}^{512}$ using a vision transformer architecture. This includes patch embedding ($32 \times 32$), positional encoding, multi-head self-attention, and a class token aggregation mechanism. Embeddings are subsequently $L_2$-normalized to lie on the unit hypersphere, enabling cosine similarity to serve as the primary similarity metric:

\begin{multline}
\mathbf{e}_i = \text{L2\_normalize}(\text{ViT}_{\text{CLIP}}(\mathbf{I}_i)), \\
\quad \text{sim}(\mathbf{e}_i, \mathbf{e}_j) = \cos(\theta_{ij}) = \mathbf{e}_i \cdot \mathbf{e}_j
\end{multline}

All reference embeddings are stored in a vector database implemented using Facebook AI Similarity Search (FAISS), employing an exact inner product index (\texttt{IndexFlatIP}). Each entry is paired with a category label \(y_i\) taking values \textit{healthy}, \textit{black-rot}, \textit{rust}, \textit{scab}, and associated metadata (e.g., image URLs, index references).

Each entry is paired with a category label
\(y_i\in\{\text{healthy},\allowbreak\ \text{black-rot},\allowbreak\ \text{rust},\allowbreak\ \text{scab}\}\)
and associated metadata (e.g., image URLs, index references).

At inference time, the system performs $k$-nearest neighbor retrieval. Given a query image $\mathbf{I}_q$, we compute its embedding $\mathbf{e}_q$ and retrieve the top-$k$ most similar images:
\begin{equation}
R_k = \arg\max_k \{\cos(\mathbf{e}_q, \mathbf{e}_i) \mid i \in [1, N]\}
\end{equation}
Each retrieved item is assigned a similarity score $s_i = \cos(\mathbf{e}_q, \mathbf{e}_i)$.

To classify the query image, the system applies a weighted voting mechanism in which the confidence for each category $c$ is computed by normalizing the similarity-weighted votes from the retrieved examples:
\begin{equation}
\text{conf}(c) = \frac{\sum_{i \in R_k, y_i = c} s_i}{\sum_{j \in R_k} s_j}
\end{equation}
This produces interpretable confidence scores that reflect both the quantity and quality of visual evidence for each class.

Implementation details include batch processing for efficiency, robust error handling (e.g., corrupted images, missing URLs), and database persistence using FAISS binary formats and Python serialization. The flat index provides exact search with $O(N)$ query complexity, sufficient for moderate dataset sizes. For larger-scale deployment, approximate nearest neighbor indexing may be integrated.

When queried, the Image-RAG system returns a ranked list of candidate categories along with normalized confidence scores, e.g.:
\begin{verbatim}
[
  {"category": "scab", "confidence": 0.5005},
  {"category": "healthy", "confidence": 0.3996},
  {"category": "rust", "confidence": 0.0999}
]
\end{verbatim}

This interpretability and modularity make Image-RAG a natural supplement to the broader Agentic AI classification pipeline.

All source code, algorithm and data is publicly available at https://github.com/Applied-AI-Research-Lab/Orchestrator-Agent-Trust

\section{Results}
\label{sec2}

\subsection{Zero-Shot Agentic Classification with Confidence-Aware Orchestration}
\label{sec21}

In the first experimental configuration, we evaluated the baseline performance of a modular Agentic AI system in a zero-shot classification setting. Two general-purpose multimodal agents, GPT-4o and Qwen-2.5-VL, were deployed without any task-specific fine-tuning. Each agent received an image prompt and independently produced a classification label, a natural language rationale, and a normalized confidence score. These outputs were then passed to a non-visual reasoning orchestrator (o3-mini-2025-01-31), which performed structured comparative reasoning to synthesize a final classification decision. The orchestrator operated under a confidence-aware strategy, weighting agent responses based on reported confidence and justification coherence, without any external retrieval or trust calibration mechanisms. This configuration establishes a foundational benchmark to assess model reliability, ensemble effectiveness, and overconfidence dynamics in open-domain zero-shot inference.

\begin{figure*}[!htbp]
\centering
\includegraphics[width=\textwidth]{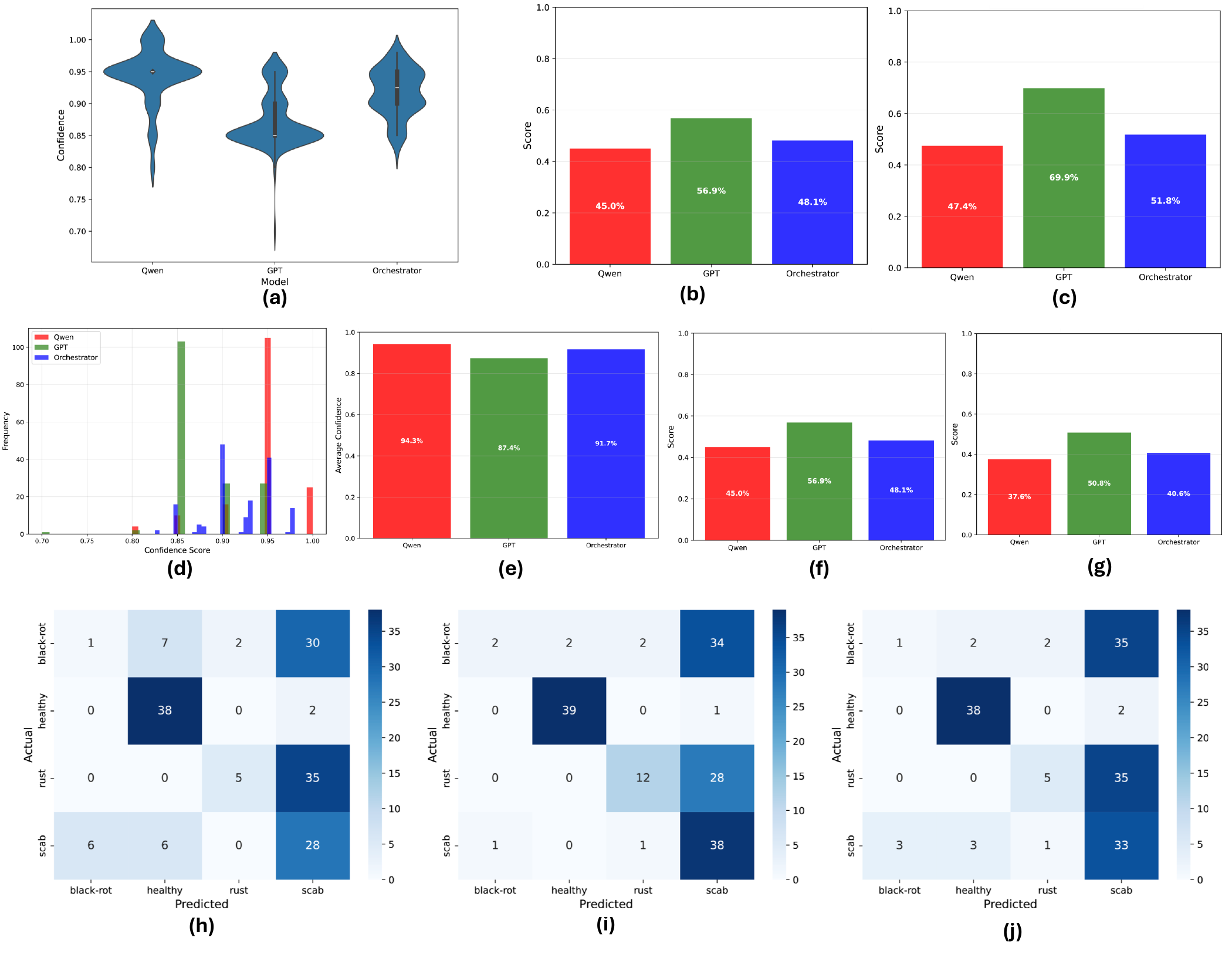}
\caption{\textbf{Experiment I – Zero-Shot Performance.} 
(a) Violin plot of model confidence distributions across all classes for Qwen-2.5-VL, GPT-4o, and the orchestrator in the zero-shot configuration (Experiment I). (b) Top-1 classification accuracy comparison of Qwen, GPT, and the orchestrator in Experiment I. (c) Weighted precision scores of all agents in the zero-shot setting. (d) Histogram of raw confidence scores reported by Qwen, GPT, and the orchestrator. (e) Mean confidence scores per agent, aggregated across the entire test set. (f) Weighted recall for each agent in zero-shot inference. (g) Weighted F1-scores for Qwen, GPT, and the orchestrator, highlighting performance balance. (h) Confusion matrix for Qwen-2.5-VL predictions on the evaluation set. (i) Confusion matrix for GPT-4o predictions, revealing inter-class confusion patterns. (j) Confusion matrix for the final orchestrator decision outcomes based on trust-aware arbitration.}
\label{fig:exp1}
\end{figure*}

The performance metrics of the agents and orchestrator are summarized in Table~\ref{tab:performance_metrics}. GPT-4o achieved the highest zero-shot accuracy at 56.88\%, followed by the orchestrator at 48.13\% and Qwen at 45.00\%. Confidence score distributions (Fig.~\ref{fig:exp1}a) revealed significant disparities in self-reported certainty, with Qwen averaging 94.3\% confidence versus GPT-4o's 87.4\% (Table~\ref{tab:app_exp1_confidence}). However, this confidence was not reliably predictive of correctness; overconfident misclassifications were common, particularly for Qwen, contributing to a higher OCR. The orchestrator’s comparative logic led to a marginal calibration improvement, though the resulting accuracy remained below 50\%.

Figure~\ref{fig:exp1}b reports top-1 classification accuracy for each agent, where GPT-4o clearly outperforms the others. Weighted precision scores are illustrated in Fig.\ref{fig:exp1}c, and raw confidence distributions are presented in Fig.\ref{fig:exp1}d. Average confidence levels per model are plotted in Fig.\ref{fig:exp1}e, with Qwen showing systematic overconfidence. Recall and F1 scores are detailed in Figs.\ref{fig:exp1}f and~\ref{fig:exp1}g, respectively, showing GPT-4o leading across metrics.

Confusion matrices in Figs.~\ref{fig:exp1}h, ~\ref{fig:exp1}i, and ~\ref{fig:exp1}j display inter-class prediction patterns for Qwen, GPT, and the orchestrator, respectively. Misclassification frequently occurred between visually similar disease classes (e.g., black-rot and rust), highlighting the inherent difficulty of this task for zero-shot models.

These findings reveal that while zero-shot multimodal agents can produce fluent and confident predictions, their self-reported confidence often fails to align with empirical performance. Qwen’s overconfidence produced high OCR values and unstable outcomes, while GPT-4o demonstrated slightly better alignment between confidence and correctness. The orchestrator, despite lacking visual input, added interpretability and decision consistency through structured reasoning. However, the marginal performance gains and misalignment of confidence and accuracy underscore the need for deeper trust calibration. These limitations served as motivation for the next experimental stages introducing supervised fine-tuning (Experiment II) and a full trust-aware orchestration with re-evaluation mechanisms (Experiment III).

\begin{table*}[t!]
\centering
\caption{
Performance metrics (accuracy, weighted and macro precision, recall, and F1) for each agent across three experimental configurations. All metrics for each experiment are shown in the same row.
}
\renewcommand{\arraystretch}{1.2}
\begin{tabularx}{\textwidth}{p{2.7cm} X X X}
\toprule
\textbf{Experiment} & \textbf{Qwen} & \textbf{GPT} & \textbf{Orchestrator} \\
\midrule
\textbf{I. Zero-Shot + Confidence-Aware}
& \small
\begin{tabular}[t]{@{}l@{}}
Accuracy: 0.4500\\
Precision$_w$: 0.4742\\
Recall$_w$: 0.4500\\
F1-score$_w$: 0.3763\\
Precision$_m$: 0.4742\\
Recall$_m$: 0.4500\\
F1-score$_m$: 0.3763
\end{tabular}
& \small
\begin{tabular}[t]{@{}l@{}}
Accuracy: 0.5688\\
Precision$_w$: 0.6985\\
Recall$_w$: 0.5688\\
F1-score$_w$: 0.5078\\
Precision$_m$: 0.6985\\
Recall$_m$: 0.5688\\
F1-score$_m$: 0.5078
\end{tabular}
& \small
\begin{tabular}[t]{@{}l@{}}
Accuracy: 0.4813\\
Precision$_w$: 0.5183\\
Recall$_w$: 0.4813\\
F1-score$_w$: 0.4062\\
Precision$_m$: 0.5183\\
Recall$_m$: 0.4813\\
F1-score$_m$: 0.4062
\end{tabular}
\\
\addlinespace
\textbf{II. Few-Shot + Confidence-Aware}
& \small
\begin{tabular}[t]{@{}l@{}}
Accuracy: 0.9563\\
Precision$_w$: 0.9603\\
Recall$_w$: 0.9563\\
F1-score$_w$: 0.9558\\
Precision$_m$: 0.9603\\
Recall$_m$: 0.9563\\
F1-score$_m$: 0.9558
\end{tabular}
& \small
\begin{tabular}[t]{@{}l@{}}
Accuracy: 0.9813\\
Precision$_w$: 0.9817\\
Recall$_w$: 0.9813\\
F1-score$_w$: 0.9812\\
Precision$_m$: 0.9817\\
Recall$_m$: 0.9813\\
F1-score$_m$: 0.9812
\end{tabular}
& \small
\begin{tabular}[t]{@{}l@{}}
Accuracy: 0.9750\\
Precision$_w$: 0.9765\\
Recall$_w$: 0.9750\\
F1-score$_w$: 0.9747\\
Precision$_m$: 0.9765\\
Recall$_m$: 0.9750\\
F1-score$_m$: 0.9747
\end{tabular}
\\
\addlinespace
\textbf{III. Zero-Shot + Trust-Aware + RAG}
& \small
\begin{tabular}[t]{@{}l@{}}
Accuracy: 0.7313\\
Precision$_w$: 0.7526\\
Recall$_w$: 0.7313\\
F1-score$_w$: 0.7292\\
Precision$_m$: 0.7526\\
Recall$_m$: 0.7313\\
F1-score$_m$: 0.7292
\end{tabular}
& \small
\begin{tabular}[t]{@{}l@{}}
Accuracy: 0.8750\\
Precision$_w$: 0.8898\\
Recall$_w$: 0.8750\\
F1-score$_w$: 0.8690\\
Precision$_m$: 0.8898\\
Recall$_m$: 0.8750\\
F1-score$_m$: 0.8690
\end{tabular}
& \small
\begin{tabular}[t]{@{}l@{}}
Accuracy: 0.8563\\
Precision$_w$: 0.8591\\
Recall$_w$: 0.8563\\
F1-score$_w$: 0.8555\\
Precision$_m$: 0.8591\\
Recall$_m$: 0.8563\\
F1-score$_m$: 0.8555
\end{tabular}
\\
\bottomrule
\end{tabularx}
\label{tab:performance_metrics}

\footnotesize
Weighted (subscript $w$) and macro-averaged (subscript $m$) precision, recall, and F1 scores are reported.

\end{table*}

\subsection{Fine-Tuned Agentic Models with Confidence-Aware Orchestration}
\label{sec22}

To evaluate whether supervised adaptation improves model reliability and ensemble synergy, we conducted a second experiment in which both GPT-4o and Qwen-2.5-VL were fine-tuned on the apple disease classification dataset. The overall setup mirrored Experiment I, retaining identical multimodal inputs, test prompts, and classification objectives. However, the zero-shot agents were replaced with domain-adapted variants, each fine-tuned on a labeled training set of 512 samples and validated on 128 additional samples. Importantly, the orchestrator (o3-mini-2025-01-31) remained unchanged and continued to perform confidence-aware arbitration, relying solely on agent-generated confidence scores and textual justifications to make final decisions.

This design tests whether fine-tuning alone, without any updates to the orchestration logic or external retrieval modules, could significantly enhance prediction accuracy, calibration, and inter-agent agreement. Table~\ref{tab:performance_metrics} presents the updated metrics: GPT-4o achieved an accuracy of 98.13\%, while Qwen-2.5-VL reached 95.63\%. The orchestrator, despite lacking direct visual input, attained 97.50\% accuracy, underscoring the impact of improved agent outputs on system-wide arbitration. Weighted and macro-averaged precision, recall, and F1 scores exceeded 95\% for all models, indicating a substantial leap from the zero-shot performance baseline.

\begin{figure*}[!htbp]
\centering
\includegraphics[width=\textwidth]{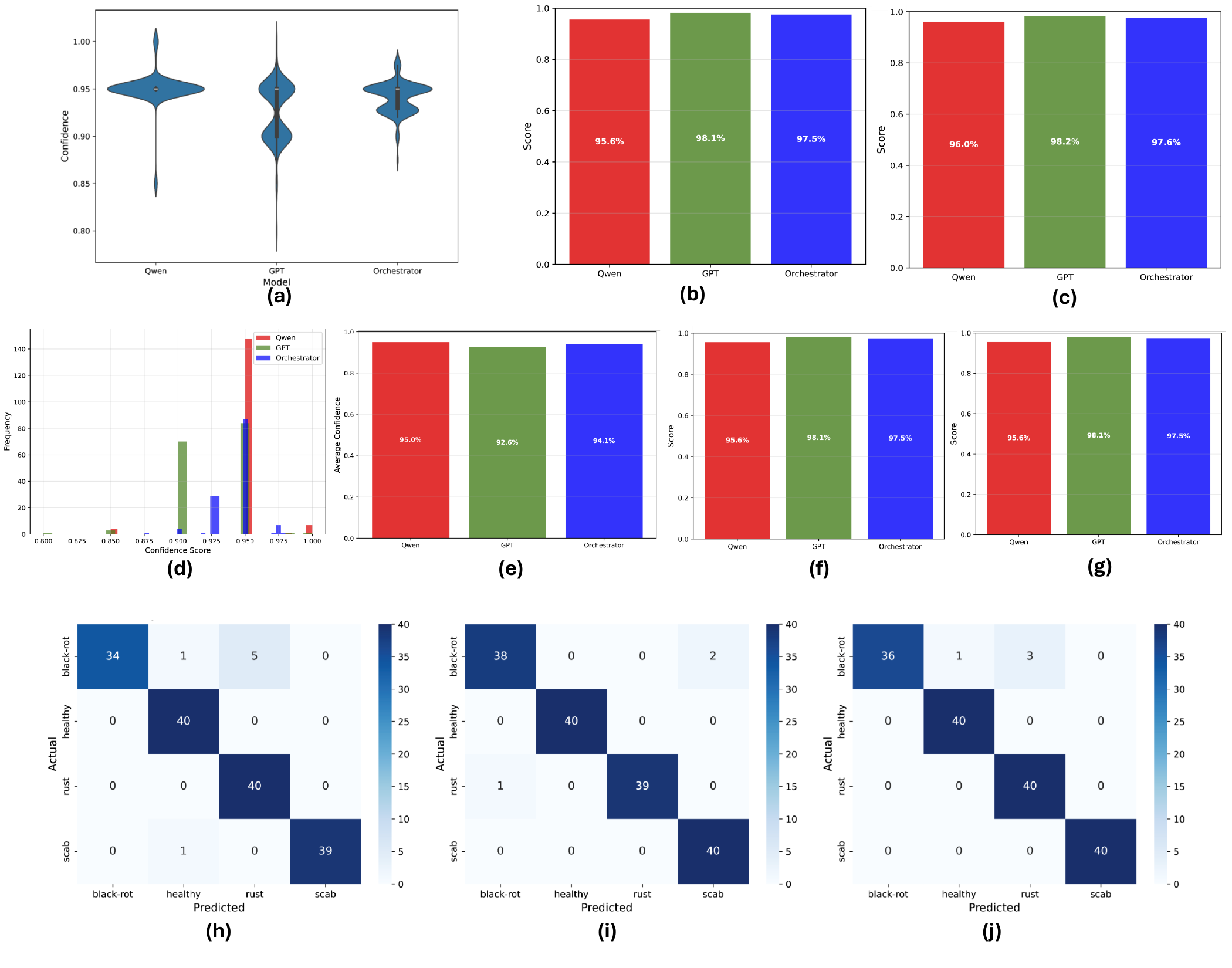}
\caption{\textbf{Experiment II – Fine-Tuned Agents.} 
(a) Violin plot of confidence distributions across all classes for Qwen-2.5-VL, GPT-4o, and the orchestrator following supervised fine-tuning. 
(b) Top-1 classification accuracy comparison after agent fine-tuning. 
(c) Weighted precision scores across the three agents, showing enhanced discriminative performance. 
(d) Histogram of raw confidence scores across all predictions, reflecting sharper calibration post fine-tuning. 
(e) Mean confidence scores per agent, showing convergence in self-reported certainty. 
(f) Weighted recall scores for Qwen, GPT, and the orchestrator on the test set. 
(g) Weighted F1-scores highlighting overall performance balance improvements. 
(h) Confusion matrix for Qwen-2.5-VL showing improved inter-class discrimination. 
(i) Confusion matrix for GPT-4o illustrating reduced misclassification frequency. 
(j) Confusion matrix for the orchestrator's final decisions, demonstrating stability in arbitration after fine-tuning.}
\label{fig:exp2}
\end{figure*}

Figure~\ref{fig:exp2}a shows violin plots of agent confidence distributions after fine-tuning. Compared to the zero-shot setting, these distributions are narrower and more centered, suggesting improved calibration. Figure~\ref{fig:exp2}b reports top-1 accuracy, while Fig.\ref{fig:exp2}c highlights gains in weighted precision across agents. Notably, the confidence histograms in Fig.\ref{fig:exp2}d exhibit reduced overconfident noise, and the average confidence values in Fig.~\ref{fig:exp2}e show a modest increase (e.g., GPT-4o rose from 87.4\% to 92.58\%), indicating better alignment between certainty and correctness (Table~\ref{tab:app_exp2_confidence}).

Recall and F1 improvements are presented in Figs.~\ref{fig:exp2}f and \ref{fig:exp2}g, respectively. The F1 score for GPT-4o increased from 50.78\% to 98.12\%, and for Qwen from 37.63\% to 95.58\%. These gains reveal that fine-tuning not only enhances correct predictions but also reduces false positives and negatives. Confusion matrices in Figs.~\ref{fig:exp2}h–\ref{fig:exp2}j show sharply improved class-level discrimination. Notably, categories like black-rot and rust, which previously showed misclassification overlap, now exhibit strong diagonal dominance.

The orchestrator benefited substantially from the improved agent outputs. Since its decision logic depends on comparative evaluation of confidence-aligned explanations, the increase in calibration quality directly translates to more reliable ensemble predictions. Interestingly, although the orchestrator does not receive image inputs, its final predictions align well with the dominant and better-calibrated agent in most cases.

The results confirm that supervised adaptation significantly enhances the classification performance of general-purpose vision-language agents within the agentic AI framework. After fine-tuning on the domain-specific apple disease dataset, both GPT-4o and Qwen-2.5-VL showed notable gains across accuracy, precision, recall, and F1 scores. As shown in Table~\ref{tab:performance_metrics}, GPT-4o achieved 98.13\% accuracy and Qwen 95.63\%, while the orchestrator despite lacking direct image access reached 97.50\% accuracy. This highlights how confidence-aware arbitration benefits from well-calibrated agent responses, with the orchestrator’s performance reflecting the value of alignment between self-reported confidence and correctness.

However, the experiment also revealed a key limitation: when both fine-tuned agents confidently agree on an incorrect prediction, the orchestrator lacks a dissenting signal and cannot intervene. This consensus failure mode exposes the risk of overreliance on internal confidence and justifications alone. To address such blind spots, the next experimental framework integrates external trust metrics and retrieval-based verification. These enhancements aim to assess not only what agents predict, but also how trustworthy and evidence-grounded those predictions are especially under ambiguity or in high-stakes settings.

\subsection{Trust-Aware Orchestration with RAG and Re-Evaluation Loops}
\label{sec23}

To overcome the reliability ceiling observed in previous settings, the third experiment introduces a trust-aware orchestration framework augmented with a RAG  pipeline. Unlike Experiment II, this configuration retains the zero-shot GPT-4o and Qwen-2.5-VL agents to investigate whether reasoning and calibration improvements can be realized without task-specific fine-tuning. Enhancements to the orchestrator include: (i) a multi-metric trust scoring module incorporating ECE, OCR, and CG, and (ii) a dynamic Re-Evaluation Loop, triggered when agent trust scores fall below a learned threshold.

In this setup, low-trust predictions trigger the orchestrator to initiate a re-evaluation loop, prompting agents with their prior responses and augmenting the input with semantic retrievals from Image-RAG. The retrieval module uses CLIP (ViT-B/32) to encode a curated set of class-representative reference images into 512-dimensional embeddings, which are $L_2$-normalized and stored in a FAISS vector database using an exact inner product index (\texttt{IndexFlatIP}). Each reference embedding is associated with a disease category label $y_i \in {\text{healthy}, \text{black-rot}, \text{rust}, \text{scab}}$ and enriched with metadata such as image URLs and textual definitions. During re-evaluation, the agents receive the top-$k$ most similar reference examples based on cosine similarity, integrated into an updated prompt. This guides them to revise or reaffirm their prediction with additional visual and semantic context. The orchestrator then assesses the updated responses using trust metrics (ECE, OCR, CG) and finalizes the decision based on the most coherent and trustworthy agent output.

\begin{figure*}[!htbp]
\centering
\includegraphics[width=\textwidth]{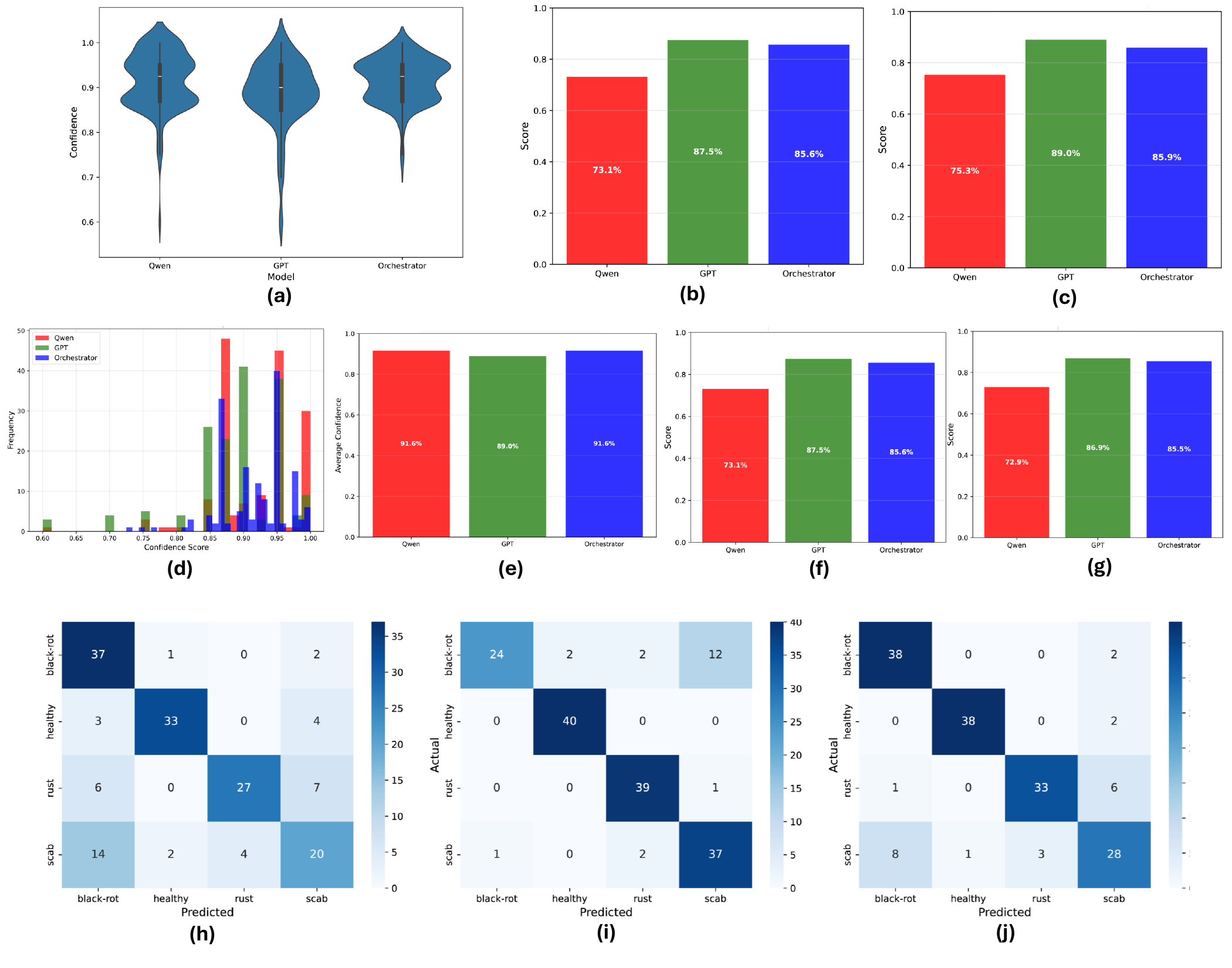}
\caption{\textbf{Experiment III – Trust-Aware Orchestration with RAG and Re-Evaluation Loops.} 
(a) Violin plot depicting confidence distributions across all classes for Qwen-2.5-VL, GPT-4o, and the orchestrator in the trust-aware setup with retrieval augmentation and re-evaluation. (b) Accuracy comparison across the three agents under trust-calibrated arbitration. (c) Weighted precision scores for all models following trust-aware reasoning. (d) Histogram of raw confidence outputs from each agent after trust score filtering. (e) Mean confidence values per agent post re-evaluation, showing enhanced calibration and reduced overconfidence. (f) Weighted recall metrics across all models in the final pipeline. (g) Weighted F1-scores reflecting the balance of precision and recall under trust-informed decision-making. (h) Confusion matrix for Qwen-2.5-VL predictions after Image-RAG integration and re-evaluation. (i) Confusion matrix for GPT-4o responses within the trust-aware system. (j) Final decision confusion matrix of the orchestrator, highlighting improvements in accuracy and reduced inter-class confusion due to trust filtering and context-grounded retrieval.}
\label{fig:exp3}
\end{figure*}

\textbf{Performance:} As summarized in Table~\ref{tab:performance_metrics}, GPT-4o achieved 87.50\% accuracy and Qwen reached 73.13\%, both significantly outperforming their initial zero-shot baselines in Experiment I. The orchestrator attained an accuracy of 85.63\%, demonstrating that trust-aware orchestration, combined with retrieval-based contextualization, can close a large portion of the performance gap typically addressed via fine-tuning. Subfigure~\ref{fig:exp3}b shows the accuracy improvements, while subfigures~\ref{fig:exp3}c,~\ref{fig:exp3}f, and~\ref{fig:exp3}g illustrate the precision, recall, and F1-score gains, respectively.

\textbf{Calibration:} As shown in Fig.~\ref{fig:exp3}a, the violin plot of confidence distributions revealed a reduction in overconfidence for both agents. Mean confidence scores (Fig.~\ref{fig:exp3}e) slightly decreased compared to Experiment II, suggesting more cautious, better-calibrated outputs. The orchestrator’s histogram (Fig.~\ref{fig:exp3}d) reflects this calibrated behavior, with lower variance and reduced extremity in self-reported confidence (Table~\ref{tab:app_exp3_confidence_postrag}).

\textbf{Qualitative Gains:} Confusion matrices shown in Figs.~\ref{fig:exp3}h–j provide insight into class-specific performance. RAG and trust-based re-evaluation mitigated frequent misclassifications seen in Experiment I especially in visually ambiguous classes such as black-rot and rust by incorporating external visual-textual anchors. Notably, GPT-4o showed strong diagonal dominance in its matrix, and the orchestrator successfully avoided the propagation of low-trust predictions even when both agents initially erred.

\textbf{Trust profiling:} From the profiling, several insights emerge regarding agent reliability under zero-shot conditions. Importantly, both models exhibit suboptimal calibration and trustworthiness metrics overall (Table~\ref{tab:app_exp3_trust}) an expected outcome given that they were not fine-tuned on the task-specific dataset. Nonetheless, relative differences offer valuable guidance for orchestration design:
\begin{itemize}
    \item GPT demonstrates superior calibration, with a lower ECE (ECE = 0.293 vs. 0.453 for Qwen) and a substantially higher Confidence–Correctness Correlation (CCC = 0.361, $p < 0.0001$), suggesting that its confidence estimates are more aligned with empirical correctness and thus more actionable in trust-aware orchestration.
    \item GPT exhibits greater reliability under uncertainty, evidenced by a lower Overconfidence Rate (OCR = 0.416 vs. 0.508) and fewer high-confidence errors (213 vs. 260), indicating a reduced risk of confidently incorrect predictions a critical feature for high-stakes decision environments.
    \item Qwen displays overconfidence and poor discrimination, reporting the highest average confidence (0.945) but a minimal Confidence Gap (CG = 0.009), meaning it struggles to differentially calibrate its confidence for correct versus incorrect predictions. This weakens its applicability in systems that depend on confidence signals for agent weighting or override logic.
\end{itemize}
These findings enabled the orchestrator to weight GPT's predictions more heavily during ensemble decision-making, thereby improving the robustness and accuracy of the overall Agentic AI system.

\textbf{Disagreement Analysis and Trust Arbitration:} An analysis of disagreements between agents and system components provides further insight:
In 12.5\% of cases, the GPT agent refused the orchestrator's recommendation for re-evaluation, returning the same prediction as before. However, in only 3 out of those 20 cases did this result in a correct prediction.
The same pattern, but to a greater extent, was observed with Qwen: the agent ignored the re-evaluation recommendation in 29.38\% of cases, returning the same prediction. Yet only 5 out of 47 such instances led to a correct outcome.

These low correctness rates in disagreement scenarios suggest that the orchestrator, re-evaluation loop, and Image-RAG remain authoritative sources in edge cases although the effectiveness of this integration benefits from final arbitration by a meta-reasoning agent.

Further analysis of orchestrator-agent disagreements shows:

\begin{itemize}
\item Orchestrator vs. GPT: Disagreements occurred 36 times (22.5\%), with the orchestrator being correct in 16 of those cases (44.44\%).
\item Orchestrator vs. Qwen: Disagreements occurred 22 times (13.75\%), with the orchestrator being correct in 21 cases (95.45\%).
\end{itemize}

The orchestrator significantly outperforms Qwen in disagreement scenarios and even surpasses GPT nearly half the time, underscoring the value of trust-aware arbitration beyond simple majority voting.

However, the re-evaluation process is not without drawbacks. In some instances, it led to overcorrection:

\begin{itemize}
\item GPT changed correct answers to incorrect ones 3 times (1.88\%) after receiving a re-evaluation prompt.
\item Qwen did so 22 times (13.75\%).
\end{itemize}

While rare for GPT, Qwen's frequent revision of correct answers indicates a higher sensitivity to prompt influence, highlighting the need for better confidence calibration and tighter visual-textual integration.

\textbf{Interpretability and Scalability:} While the orchestrator achieved near-fine-tuned accuracy without labeled training data, residual failure cases remain, especially when both agents confidently agree on a wrong label. This scenario exposes the limits of current trust metrics to fully capture uncertainty in high-confidence false predictions. Nevertheless, the current framework provides a scalable, modular, and interpretable zero-shot system that balances generalizability and reliability.

Collectively, Experiment III highlights that zero-shot generalist agents, when embedded in trust-calibrated agentic AI systems with access to external retrieval, can achieve expert-level accuracy in image classification tasks. These findings support a broader vision of scalable agentic intelligence where trust, not tuning, becomes the key to real-world deployment.

\subsection{Comparative Performance and Ablation Analysis}
\subsubsection{Time and Calibration Performance Across Configurations}
\begin{figure*}[!htbp]
\centering
\includegraphics[width=\textwidth]{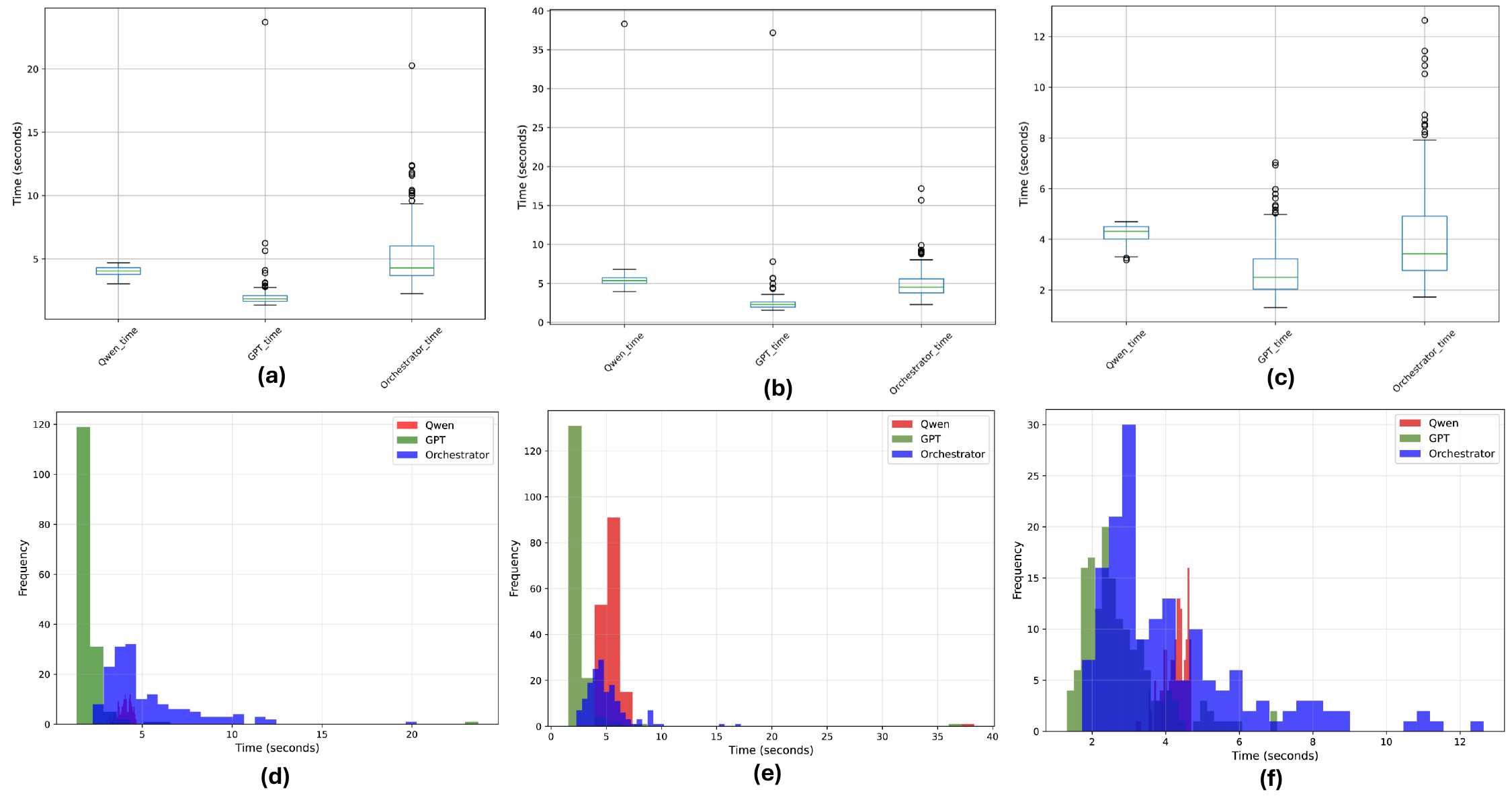}
\caption{\textbf{Time Performance Analysis across Experimental Configurations.} 
a) Boxplot showing inference time distribution per image in Experiment I (zero-shot setting).  
b) Boxplot showing inference time distribution in Experiment II (fine-tuned setting).  
c) Boxplot showing inference time distribution in Experiment III (trust-aware orchestration with RAG).  
d) Histogram of inference time frequencies for Experiment I.  
e) Histogram of inference time frequencies for Experiment II.  
f) Histogram of inference time frequencies for Experiment III.  
These visualizations highlight how orchestration strategies and model configurations affect latency, offering insight into the computational trade-offs of agentic AI systems.}
\label{fig:exp4}
\end{figure*}

Figure~\ref{fig:exp4} presents a comparative analysis of inference time across the three experimental setups. Subfigures~\ref{fig:exp4}a--c show that fine-tuned agents (Experiment II) achieve the lowest latency due to optimized internal representations, while trust-aware orchestration with retrieval (Experiment III) introduces modest time overhead from retrieval and re-evaluation cycles. Histograms in subfigures~\ref{fig:exp4}d--f reveal that Experiment III exhibits a heavier tail in inference time distribution, yet remains within real-time thresholds. While fine-tuned models yield the highest performance, they require extensive training and lack task transferability. In contrast, Experiment III achieves near-optimal accuracy (85.6\%) with only 1.3× the inference time of zero-shot baselines, offering a practical compromise for label-scarce or dynamic environments.

\subsubsection{Confidence-Accuracy Calibration and Overconfidence Mitigation}
\begin{figure*}[!htbp]
\centering
\includegraphics[width=\textwidth]{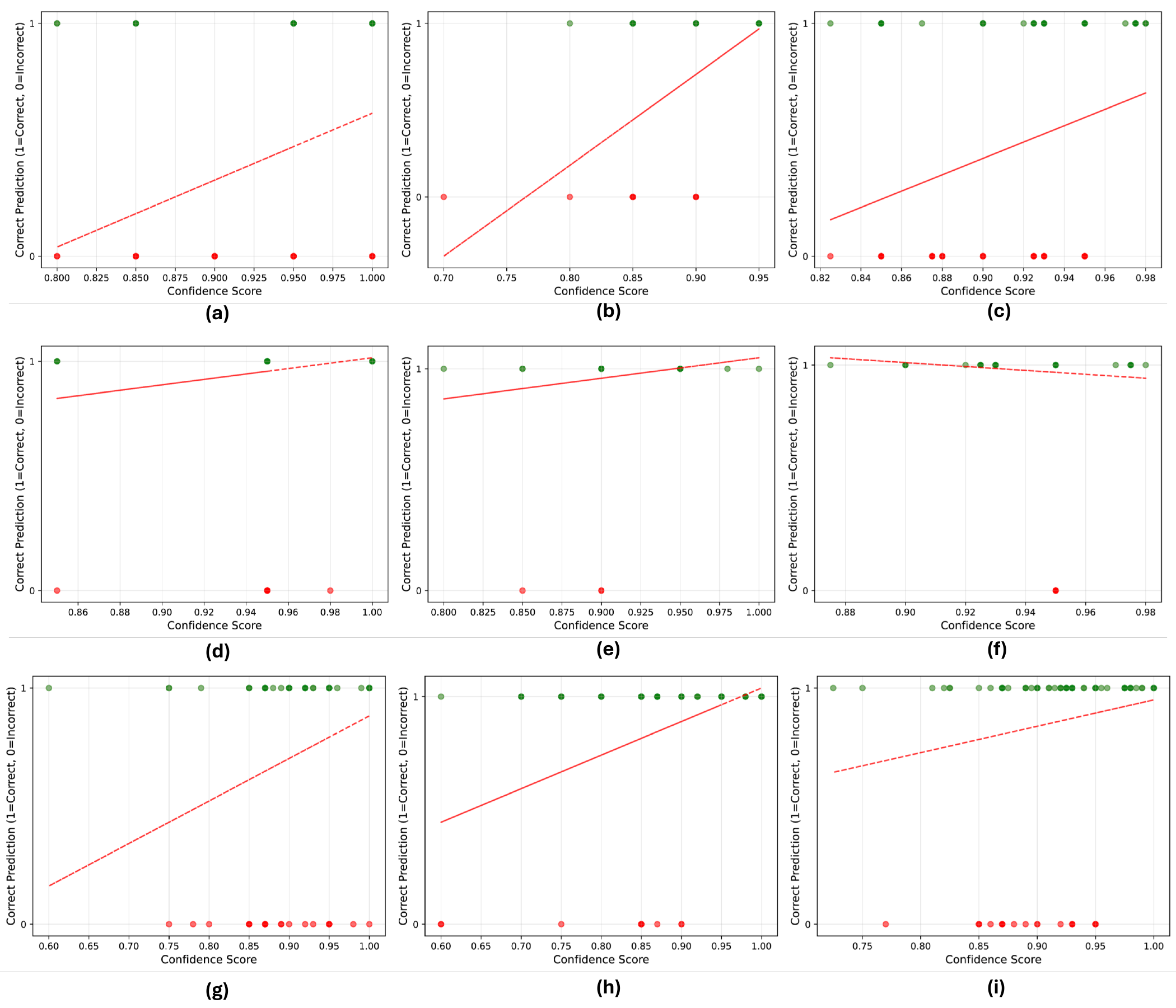}
\caption{\textbf{Confidence vs. Accuracy Calibration Analysis across Experiments.} 
a) Confidence vs. accuracy plot for Qwen-2.5-VL in Experiment I (zero-shot setting);  
b) Confidence vs. accuracy plot for GPT-4o in Experiment I;  
c) Confidence vs. accuracy plot for the orchestrator in Experiment I;  
d) Confidence vs. accuracy plot for Qwen-2.5-VL in Experiment II (fine-tuned setting);  
e) Confidence vs. accuracy plot for GPT-4o in Experiment II;  
f) Confidence vs. accuracy plot for the orchestrator in Experiment II;  
g) Confidence vs. accuracy plot for Qwen-2.5-VL in Experiment III (trust-aware orchestration with RAG);  
h) Confidence vs. accuracy plot for GPT-4o in Experiment III;  
i) Confidence vs. accuracy plot for the orchestrator in Experiment III.  
These calibration curves illustrate the alignment between predicted confidence and true accuracy across agents and experimental conditions, highlighting changes in overconfidence and calibration quality.}
\label{fig:exp5}
\end{figure*}

\begin{figure*}[!htbp]
\centering
\includegraphics[width=0.95\textwidth]{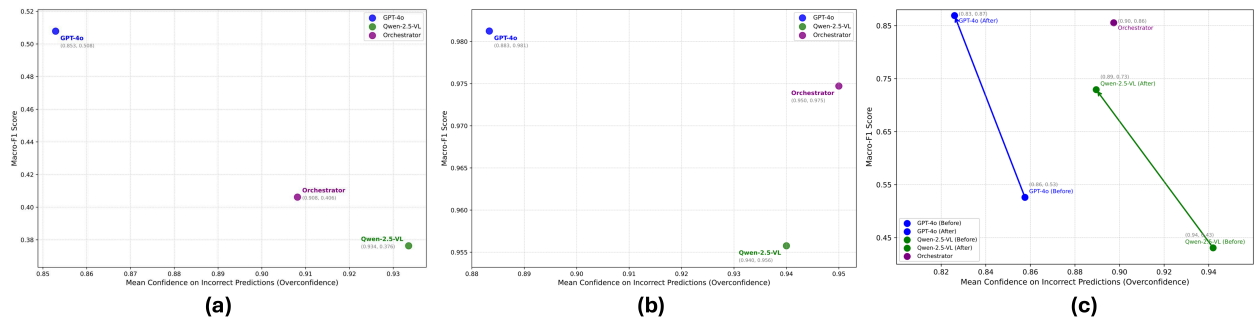}
\caption{\textbf{Agent overconfidence vs. final macro-F1.} 
(a) Experiment I: zero-shot predictions – overconfidence vs. macro-F1 score; 
(b) Experiment II: few-shot predictions – overconfidence vs. macro-F1 score; 
(c) Experiment III: trust-aware orchestration with RAG – overconfidence vs. macro-F1 score, both before and after re-evaluation loop.}
\label{fig:exp6}
\end{figure*}

Figure~\ref{fig:exp5} illustrates confidence–accuracy calibration curves for Qwen-2.5-VL, GPT-4o, and the orchestrator across all experiments. In the zero-shot setting (subfigures~\ref{fig:exp5}a--c), both agents display considerable overconfidence, with confidence often exceeding empirical accuracy. This misalignment is partially mitigated by the orchestrator, which arbitrates between agent outputs. Fine-tuning (subfigures~\ref{fig:exp5}d--f) improves calibration, especially for GPT-4o, aligning predicted confidence more closely with true correctness. Trust-aware orchestration in Experiment III (subfigures~\ref{fig:exp5}g--i) further suppresses overconfidence through calibrated re-evaluation. Figure~\ref{fig:exp6} visualizes the relationship between agent overconfidence (mean confidence on incorrect predictions) and final macro-F1, confirming that trust-augmented pipelines better align certainty with correctness.

\section{Discussion and Future Directions}
\label{sec:Discussion_and_Future_Directions}
The increasing complexity and autonomy of AI systems calls for robust, interpretable, and generalizable architectures that can reason, evaluate, and adapt in real time \cite{longo2024explainable}. This study presents a systematic exploration of a novel trust-aware agentic AI framework that blends zero-shot vision-language agents with orchestration, trust calibration, and RAG-based re-evaluation. Our three-tier experimental structure reveals key insights into how such systems can be structured for both performance and scalability while preserving explainability and adaptability.

A central insight from this study is the critical role of calibration and trust estimation in agentic AI systems. While traditional ensemble methods often aggregate agent outputs under the assumption of independent and reliable performance \cite{mackenzie2019platform, ganaie2022ensemble}, such strategies falter in real-world settings especially under zero-shot conditions where agents may exhibit systematic overconfidence or miscalibration \cite{frei2019ensemble, ojha2025navigating}. In these contexts, naïvely trusting self-reported confidence scores can lead to compounding errors, particularly in scenarios that demand high reliability and interpretability. 

To address this, we adopt a trust-aware orchestration strategy that incorporates metrics such as ECE, OCR, and CG, enabling the orchestrator to quantitatively assess the alignment between confidence and correctness, as well as the consistency of reasoning under varied prompt formulations. This shift from purely accuracy-driven aggregation to trust-calibrated decision fusion reflects a broader movement in AI system design toward epistemic robustness and risk-aware reasoning. Prior research in LLM-based decision support, autonomous robotics, and human-AI collaboration has underscored the limitations of relying on uncalibrated model outputs, and has proposed various trust modeling frameworks that incorporate self-assessment, uncertainty quantification, or post hoc calibration techniques. 

Our findings affirm that integrating trust metrics directly into orchestration logic significantly improves both accuracy and reliability. By down-weighting overconfident yet incorrect predictions and triggering re-evaluation when inconsistency or low trust is detected, the system becomes more resilient to epistemic failures. This capability is especially crucial in high-stakes domains such as medical diagnostics, autonomous driving, and scientific discovery, where misjudged confidence can lead to misinformed actions with costly or irreversible consequences. Rather than treating trust as an external interpretability add-on, our framework embeds trust evaluation as a first-class component of agentic reasoning, aligning with emerging paradigms in trust-centric AI governance and human-AI alignment.

Our findings also show that the method of orchestration itself profoundly affects system performance. In the baseline configuration (Experiment I), the orchestrator made decisions solely based on self-reported confidence, leading to moderate accuracy (48.13\%) and significant overconfidence in incorrect predictions. In contrast, the trust-aware orchestrator (Experiment III) reached an accuracy of 85.63\%, despite using zero-shot agents, highlighting the orchestration logic as a performance amplifier. Moreover, when compared to Experiment II, which involved computationally expensive supervised fine-tuning of agents (achieving 97.50\% accuracy), the trust-aware method captured over 77\% of the possible gain while avoiding the need for model retraining. This demonstrates that a well-calibrated orchestration mechanism can partially substitute for domain adaptation when retraining is not feasible.

Traditional ensemble learning methods, such as majority voting \cite{singh2019rna, yang2023survey}, mean averaging or confidence-weighted fusion \cite{brown2017ensemble}, operate under the assumption that model outputs are statistically independent and equally reliable. While effective in low-noise environments or when models are homogeneously calibrated, these techniques struggle in heterogeneous, zero-shot, or high-uncertainty scenarios where agent predictions may be misaligned or systematically overconfident \cite{jahan2025uncertainty, he2023survey}. In contrast, our trust-aware orchestrator does not merely aggregate predictions it actively evaluates each agent’s trustworthiness using multi-dimensional metrics and selectively down-weights or discards predictions that exhibit poor calibration, inconsistency, or unjustified confidence. This approach aligns more closely with emerging frameworks in agentic AI, where orchestration involves structured reasoning across multiple autonomous agents with varying competencies.

Prior work on dynamic task decomposition \cite{flores2025structured, gao2024large}, agent delegation \cite{fernandez2022delegation, pataranutaporn2021ai}, and modular reasoning \cite{lu2025fine, odobesku2025agent} has shown that intelligent coordination across agents can outperform flat ensembles, particularly when agents contribute distinct skills or modalities \cite{liu2025embodied}. However, many of these systems rely on rule-based or deterministic coordination logic and lack mechanisms for trust-based arbitration or reflective re-evaluation \cite{kermansaravi2025ai, li2025investigation}. Our orchestrator extends this space by integrating retrieval-augmented prompts and dynamic trust profiles, enabling recursive decision correction based on evidence-grounded feedback. Compared to static ensembles, this architecture enables real-time reasoning under uncertainty, improves robustness to adversarial disagreement, and supports scalable integration of new agents without retraining. Such orchestrator logic is increasingly critical as agent ecosystems grow in complexity and move toward plug-and-play, open-world operation.

An additional strength of our framework lies in the re-evaluation loop powered by RAG. This feedback mechanism enables AI agents to reflect on their prior decisions in light of retrieved evidence from a vision-language knowledge base. By structuring the retrieved information around interpretable class definitions and visual prototypes, the system compensates for hallucination or semantic ambiguity common in zero-shot models. Our ablation analysis shows that although this loop was triggered in 100\% of instances due to low trust profiling scores, disagreements between agents and system components reveal that GPT and Qwen often ignored re-evaluation prompts 12.5\% and 29.38\% of cases, respectively but this rarely led to correct predictions (only 3 out of 20 for GPT and 5 out of 47 for Qwen). These low success rates highlight the importance of the orchestrator, re-evaluation loop, and Image-RAG as authoritative sources, with optimal outcomes achieved through meta-reasoning arbitration. The practical implication is that retrieval-based grounding acts as an auxiliary supervision signal, enabling improvement without manual annotation or gradient updates.

While RAG has shown considerable promise in enhancing large language models through external knowledge grounding especially in tasks such as medical decision support \cite{wada2025retrieval, ke2025retrieval}, document retrieval  and language alignment these systems typically lack a structured trust arbitration mechanism \cite{dong2025understand}. Most RAG implementations retrieve top-k textual or visual exemplars to refine responses, but treat the retrieval step as static and apply equal weight to all retrieved content \cite{yang2025retrieval, zhang2025leveraging, prince2024opportunities}. In contrast, our Image-RAG pipeline integrates visual retrieval with dynamic trust scoring, enabling iterative re-evaluation loops where low-confidence or conflicting predictions trigger a targeted grounding process. This allows the system to not only retrieve relevant cases but also modulate decision-making based on model reliability and retrieval quality. Existing document-grounded agents focus on improving factuality but do not incorporate agent-level trust profiling \cite{hammane2024selfrewardrag}. Moreover, few approaches fuse visual similarity with structured re-prompting guided by trust thresholds \cite{li2024survey}. Our design enhances interpretability by showing not only what evidence was retrieved but why it was considered trustworthy. This feedback loop is essential for deploying agentic AI in settings where error introspection and evidence traceability matter, and it represents a step beyond static RAG toward trust-calibrated retrieval orchestration.

From a system design perspective, our architecture prioritizes modularity and plug-and-play scalability, enabling seamless integration of agents with minimal friction. Each agent operates autonomously and can be added, removed, or updated without necessitating retraining of the orchestrator a critical property for real-world deployment where agent capabilities may evolve over time. This decoupled architecture reflects foundational principles from modular agent systems developed in domains such as edge computing and the Internet of Things, where component isolation, interoperability, and system composability are essential for dynamic, distributed environments. 

As agentic AI systems scale to incorporate dozens or even hundreds of specialized agents each with different modalities, competencies, or domain knowledge per-agent fine-tuning becomes computationally impractical and operationally rigid \cite{ma2024coevolving, zhou2024star}. Our trust-aware orchestration framework addresses this by absorbing agent heterogeneity through dynamic reliability profiling and selective arbitration, thereby supporting generalization across agents without task-specific adaptation. This design also mirrors recent advancements in user interface navigation agents and vision-language action systems, where orchestration is driven by flexible, intent-driven coordination rather than static aggregation. By separating reasoning logic from perception modules and incorporating natural language justifications, our system remains interpretable and auditable, supporting both technical transparency and human oversight two pillars necessary for scalable and trustworthy multi-agent AI ecosystems.

\subsection{Summary, Limitations, and Future Perspective}
\label{sec:summary}

\subsubsection{Summary}
This work presents an Agentic AI system that integrates trust-aware orchestration, vision-language grounding via Image-RAG, and structured re-evaluation loops. To evaluate the system's effectiveness, we conducted three distinct experiments, progressively increasing complexity and realism.

\begin{enumerate}
\item \textbf{Experiment~\hyperref[Experiment_I]{I} (Zero-Shot Orchestration)}: Two vision-language models;  Qwen-2.5-VL and GPT-4o;  were deployed in a zero-shot setting to make classification predictions, accompanied by natural language explanations and self-reported confidence scores. These outputs were evaluated by a reasoning agent (o3-mini-2025-01-31), which, based on content and trust signals, issued a final decision. This baseline Agentic AI system achieved an overall accuracy of 48.13\%.

\item \textbf{Experiment~\hyperref[Experiment_II]{II} (Fine-Tuned Agents)}: Both agents were thoroughly fine-tuned using a dedicated training set via hyperparameter optimization. The same inference and arbitration process was followed. With fine-tuned models, the Agentic AI system reached 97.50\% accuracy, establishing an upper performance bound when domain adaptation is permitted.

\item \textbf{Experiment~\hyperref[Experiment_III]{III} (Trust-Aware Agentic Framework)}: This experiment evaluates our proposed full framework, incorporating trust-aware orchestration and Image-RAG visual reasoning. Agents were prompted to make classification predictions on a hidden-labeled training set, along with self-reported confidence scores. Their responses were processed through a set of quantitative trustworthiness metrics to derive trust profiles, enabling agent-level reliability estimation independent of individual inputs.

The framework also included a state-of-the-art retrieval-augmented vision component using CLIP; each entry was paired with a category label, with reference embeddings stored in a FAISS-based vector database. At inference time, the orchestrator called upon agents to make zero-shot predictions, as in Experiment~\hyperref[Experiment_I]{I}. Their predictions, confidence scores, explanations, and trustworthiness metrics were passed to the orchestrator, which then decided whether to trust the agents' predictions or prompt a re-evaluation loop.

Agents entering the re-evaluation loop were provided with the context of their previous response and recommendations from the Image-RAG component, allowing them to revise their classification decisions. The agents' final responses were passed back to the orchestrator, which made an informed final decision. This framework, without any agent fine-tuning, achieved 85.63\% accuracy.
\end{enumerate}

These experiments clearly demonstrate the effectiveness of trust-aware orchestration in agentic AI systems, yielding up to a 77.94\% improvement over confidence-based zero-shot orchestration. Scaling such systems to hundreds of AI agents would make per-agent fine-tuning prohibitively expensive, both in time and computational cost, due to the need to identify optimal hyperparameters for each model. In contrast, our framework enables seamless integration of new agents without fine-tuning. Each agent contributes its unique capabilities in a zero-shot setting, while the trust-aware orchestrator provides the necessary context to incorporate them effectively and reliably into the broader system.

\subsubsection{Limitations}
Several limitations remain. First, although the system outperforms conventional zero-shot baselines, its accuracy still lags behind that of domain-specific, fine-tuned models. This performance gap highlights the trade-off between generalizability and task-specific optimization.

Second, in the zero-shot setting, the design and phrasing of prompts play a critical role in shaping model outputs. Despite clear and structured prompting, we observed that models;  particularly Qwen-2.5-VL;  frequently failed to follow the expected response format. To address this, we implemented a re-prompting loop with a capped number of retries, which introduces additional inference overhead and increases system complexity.

Third, the results reported in this study are based on a specific image dataset; therefore, performance on other image-based datasets may vary. However, we expect the relative trends to hold, with fine-tuned models generally outperforming zero-shot approaches.

Fourth, while we used the o3-mini model as the orchestrator for these experiments, there are several alternative models available. We specifically chose an orchestrator without visual capabilities to avoid biases introduced by its own predictions. Nonetheless, more advanced models such as o3-pro, o4-mini, or other variants with visual understanding could potentially improve orchestration performance.

\subsubsection{Future Perspective}
Advancing trust-aware agentic AI systems presents several promising directions. Incorporating orchestrators with enhanced multimodal reasoning capabilities;  such as more advanced visual-language models;  could improve the reliability and fairness of decision arbitration.

Further, optimizing prompt design and exploring adaptive prompting strategies will be essential to address current limitations in zero-shot settings, reducing the need for repeated prompting and improving compliance with response formats.

Scaling the framework to accommodate larger and more diverse populations of AI agents poses challenges related to agent management, specialization, and dynamic trust assessment. Developing methods to efficiently integrate and update agent trust profiles will be critical for maintaining system robustness.

Additionally, validating and extending the trust-aware orchestration approach across varied datasets and domains beyond image classification to include video analysis, language tasks, and multimodal reasoning;  will be important for demonstrating generalizability and broader impact.

Ultimately, embedding self-monitoring and self-improvement mechanisms within agents may enable autonomous adaptation and increased system resilience, paving the way for more robust and scalable agentic AI architectures applicable to complex real-world problems.

\section*{Acknowledgement} The publication of the article in OA mode was financially supported by HEAL-Link. This work was additionally supported in part by the National Science Foundation (NSF) and the United States Department of Agriculture (USDA), National Institute of Food and Agriculture (NIFA), through the “Artificial Intelligence (AI) Institute for Agriculture” program under Award Numbers AWD003473 and AWD004595, and USDA-NIFA Accession Number 1029004 for the project titled “Robotic Blossom Thinning with Soft Manipulators.” Additional support was provided through USDA-NIFA Grant Number 2024-67022-41788, Accession Number 1031712, under the project “Expanding UCF AI Research To Novel Agricultural Engineering Applications (PARTNER).”

\section*{Declarations}
The authors declare no conflicts of interest.

\section*{Statement on AI Writing Assistance}
ChatGPT and Grammarly were utilized to enhance grammatical accuracy and refine sentence structure; all AI-generated revisions were thoroughly reviewed and edited for relevance.

\bibliographystyle{cas-model2-names}

\bibliography{references}

\appendix
\clearpage

\section*{Appendix A. Supplementary Figures and Tables}
\renewcommand{\thefigure}{A\arabic{figure}}
\renewcommand{\theHfigure}{A\arabic{figure}}
\setcounter{figure}{0}

\begin{figure*}[!htbp]
\centering
\includegraphics[width=0.95\textwidth]{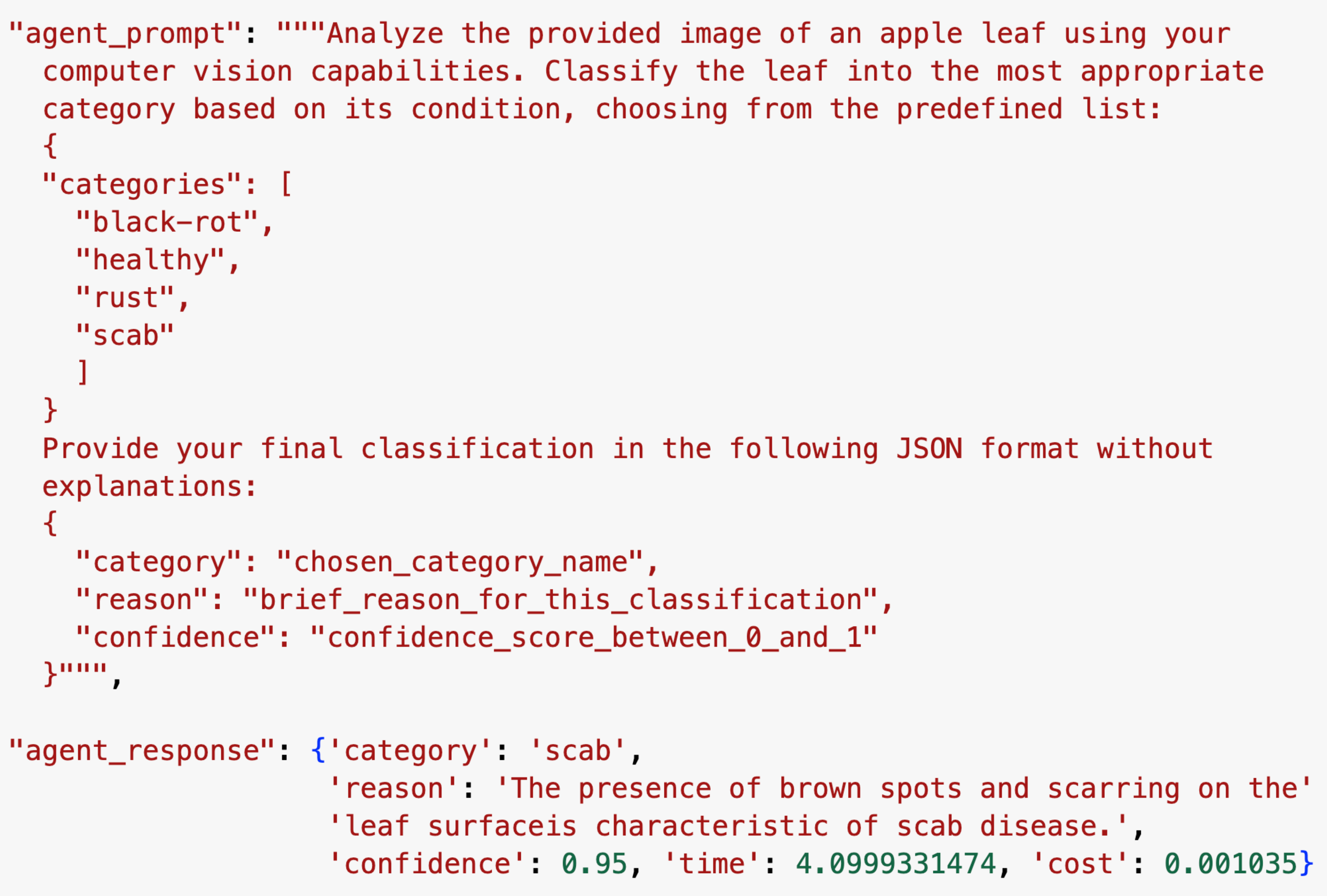}
\caption{\textbf{AI agent prompt and response for disease classification.} 
The prompt instructs a vision-language model to classify apple leaf diseases. 
The response includes the predicted category, justification, self-reported confidence, 
latency, and estimated computational cost in JSON format.}
\label{fig:appendix_agent_prompt}
\end{figure*}

\begin{figure*}[!htbp]
\centering
\includegraphics[width=0.95\textwidth]{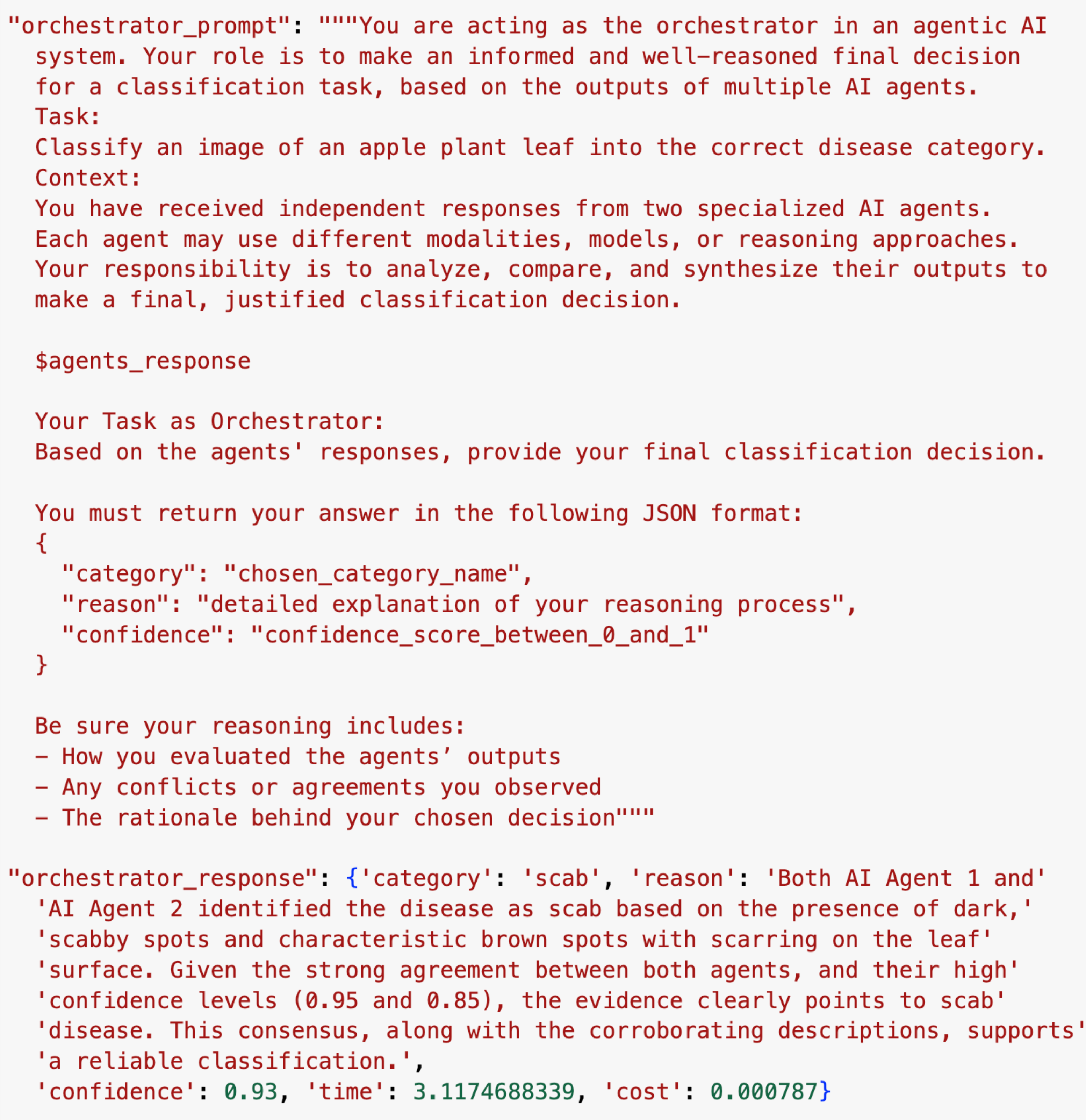}
\caption{\textbf{Orchestrator prompt and agentic response for decision arbitration.}  
The orchestrator receives outputs from multiple agents and synthesizes them into 
a single trusted decision. The JSON response includes the class, rationale, confidence, 
processing time, and cost.}
\label{fig:appendix_orchestrator_prompt}
\end{figure*}

\begin{figure*}[!htbp]
\centering
\includegraphics[width=0.95\textwidth]{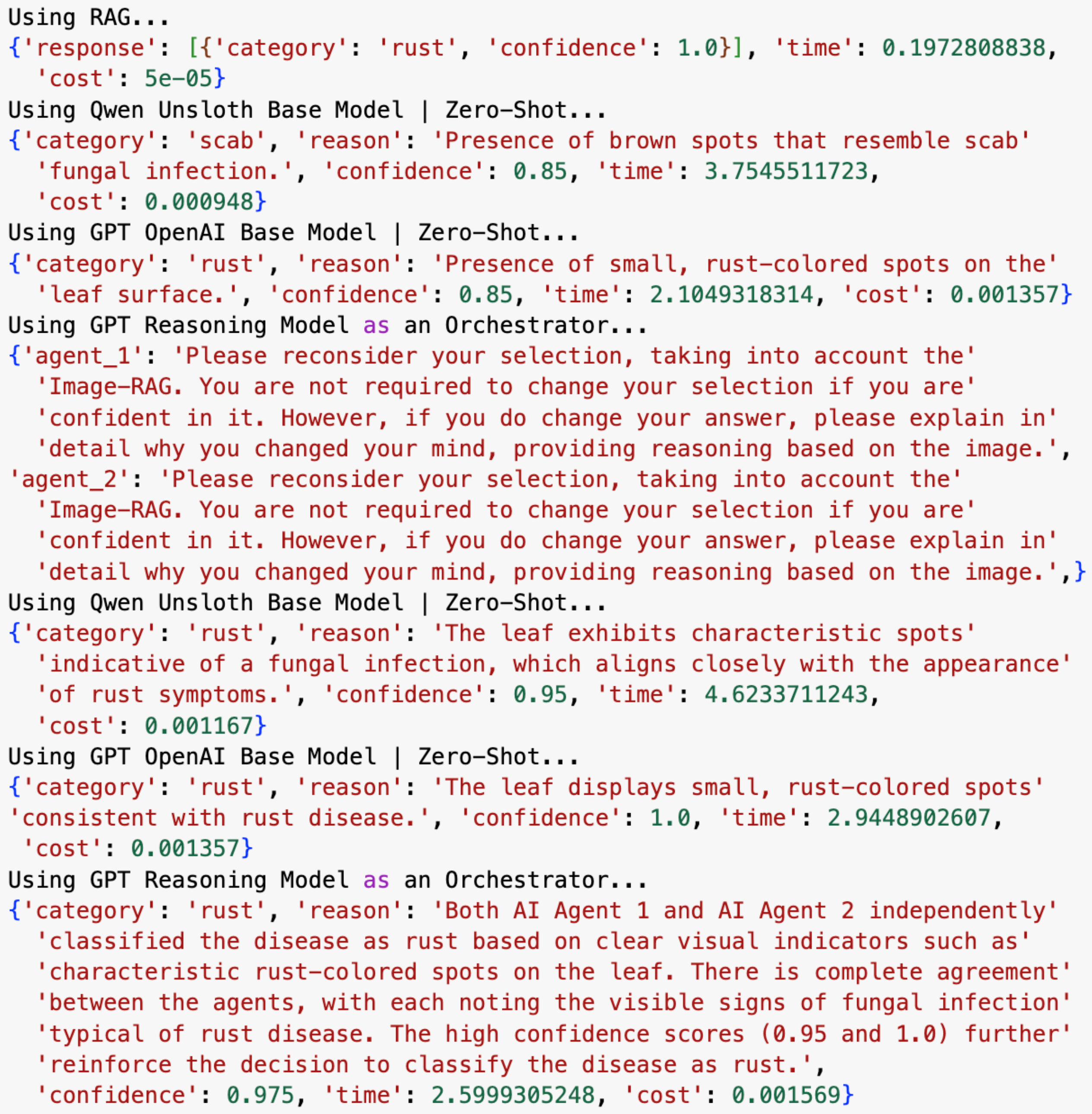}
\caption{\textbf{Experiment III: Trust-aware orchestration with RAG and re-evaluation.}  
The orchestrator triggers CLIP-based retrieval (Image-RAG) and a re-evaluation loop. 
Updated agent responses are scored by trust metrics to produce a final decision.}
\label{fig:appendix_exp3_rag}
\end{figure*}

\clearpage
\renewcommand{\thetable}{A\arabic{table}}
\setcounter{table}{0}

% ---- Table A1
\begin{table}[!htbp]
\centering
\caption{\textbf{Experiment I: Confidence score distribution for zero-shot agents and the orchestrator.}}
\begin{tabular}{lrrr}
\toprule
\textbf{Statistic} & \textbf{Qwen} & \textbf{GPT-4o} & \textbf{Orchestrator} \\
\midrule
Count & 160 & 160 & 160 \\
Mean & 0.943 & 0.874 & 0.917 \\
Std. Dev. & 0.042 & 0.042 & 0.037 \\
Min & 0.800 & 0.700 & 0.825 \\
25th Pctl & 0.950 & 0.850 & 0.900 \\
Median & 0.950 & 0.850 & 0.925 \\
75th Pctl & 0.950 & 0.900 & 0.950 \\
Max & 1.000 & 0.950 & 0.980 \\
\bottomrule
\end{tabular}
\label{tab:app_exp1_confidence}
\end{table}

% ---- Table A2
\begin{table}[!htbp]
\centering
\caption{\textbf{Experiment II: Comparison of fine-tuning settings for GPT-4o.}}
\begin{tabular}{lrrrrr}
\toprule
\textbf{Configuration} & \textbf{Epochs} & \textbf{Batch Size} & \textbf{Val. Loss} & \textbf{Duration (s)} & \textbf{Cost (USD)} \\
\midrule
GPT-4o (ResNet-50 tuned) & 10 & 16 & 0.0088 & 1778 & 47.53 \\
GPT-4o (Default settings) & 3 & 1 & 0.0617 & 1652 & 13.09 \\
\bottomrule
\end{tabular}
\label{tab:app_exp2_finetune}
\end{table}

% ---- Table A3
\begin{table}[!htbp]
\centering
\caption{\textbf{Experiment II: Confidence statistics under few-shot setting.}}
\begin{tabular}{lrrr}
\toprule
\textbf{Statistic} & \textbf{Qwen} & \textbf{GPT-4o} & \textbf{Orchestrator} \\
\midrule
Count & 160 & 160 & 160 \\
Mean & 0.950 & 0.926 & 0.941 \\
Std. Dev. & 0.019 & 0.029 & 0.016 \\
Min & 0.85 & 0.80 & 0.875 \\
25th Pctl & 0.95 & 0.90 & 0.93 \\
Median & 0.95 & 0.95 & 0.95 \\
75th Pctl & 0.95 & 0.95 & 0.95 \\
Max & 1.00 & 1.00 & 0.98 \\
\bottomrule
\end{tabular}
\label{tab:app_exp2_confidence}
\end{table}

% ---- Table A4 (wide)
\begin{table*}[!htbp]
\centering
\caption{\textbf{Experiment III: Trust metrics for zero-shot agents under trust-aware orchestration.}}
\begin{tabular}{lrrrrrrrrrrrr}
\toprule
Model & Acc. & Avg. Conf. & Conf$_\text{corr}$ & Conf$_\text{incorr}$ & CG & OCR & HCW & THC & CCC & $p$-val & ECE & CWA \\
\midrule
Qwen & 0.492 & 0.945 & 0.950 & 0.941 & 0.009 & 0.508 & 260 & 512 & 0.126 & 0.0042 & 0.453 & 0.495 \\
GPT  & 0.584 & 0.877 & 0.890 & 0.860 & 0.030 & 0.416 & 213 & 512 & 0.361 & 0.0000 & 0.293 & 0.592 \\
\bottomrule
\end{tabular}
\vspace{1mm}
{\small Note: Acc.=Accuracy; OCR=Overconfidence Ratio; HCW=High-confidence wrong; THC=Total high-confidence; CCC=Confidence-Correctness Correlation; ECE=Expected Calibration Error; CWA=Confidence-Weighted Accuracy.}
\label{tab:app_exp3_trust}
\end{table*}

% ---- Table A5
\begin{table}[!htbp]
\centering
\caption{\textbf{Experiment III: Confidence scores after trust-based re-evaluation.}}
\begin{tabular}{lrrr}
\toprule
\textbf{Statistic} & \textbf{Qwen} & \textbf{GPT-4o} & \textbf{Orchestrator} \\
\midrule
Count & 160 & 160 & 160 \\
Mean & 0.916 & 0.890 & 0.916 \\
Std. Dev. & 0.064 & 0.074 & 0.050 \\
Min & 0.60 & 0.60 & 0.725 \\
25th Pctl & 0.87 & 0.85 & 0.87 \\
Median & 0.925 & 0.90 & 0.925 \\
75th Pctl & 0.95 & 0.95 & 0.95 \\
Max & 1.00 & 1.00 & 1.00 \\
\bottomrule
\end{tabular}
\label{tab:app_exp3_confidence_postrag}
\end{table}

% ---- Table A6
\begin{table}[!htbp]
\centering
\caption{\textbf{Top 5 hyperparameter configurations ranked by validation loss.}}
\begin{tabular}{lrrrrrr}
\toprule
Rank & Trial & Val. Loss & LR & Batch & Warmup & Epochs \\
\midrule
1 & 11 & 0.000010 & $1.094 \times 10^{-5}$ & 4 & 0.0997 & 15 \\
2 & 16 & 0.000934 & $1.901 \times 10^{-5}$ & 4 & 0.0881 & 13 \\
3 & 15 & 0.000993 & $2.348 \times 10^{-5}$ & 4 & 0.0864 & 12 \\
4 & 9  & 0.001013 & $1.552 \times 10^{-4}$ & 2 & 0.0351 & 8  \\
5 & 19 & 0.001188 & $1.720 \times 10^{-5}$ & 4 & 0.0898 & 12 \\
\bottomrule
\end{tabular}
\label{tab:top_5_hyperparameter_configurations}
\end{table}

% \bibliographystyle{elsarticle-harv} 
% \bibliography{example}

\printcredits




\end{document}